\documentclass[a4paper,fleqn]{cas-dc}

\usepackage[authoryear]{natbib}

\usepackage{subcaption}
\usepackage{multirow}
\usepackage{adjustbox}
\usepackage{amsmath,amsfonts,bm}
\usepackage{alphalph}
\usepackage{amssymb}
\usepackage{pifont}
\newcommand{\xmark}{\ding{55}}%
\usepackage{comment}
\usepackage{amssymb}

\def\tsc#1{\csdef{#1}{\textsc{\lowercase{#1}}\xspace}}
\tsc{WGM}
\tsc{QE}
\tsc{EP}
\tsc{PMS}
\tsc{BEC}
\tsc{DE}

\def\eqref#1{equation~\ref{#1}}

\def\1{\bm{1}}

\newcommand{\R}{\mathbb{R}}

\newcommand{\model}{ProbGLC}

\begin{document}
\let\WriteBookmarks\relax
\def\floatpagepagefraction{1}
\def\textpagefraction{.001}

\shorttitle{\model}

\shortauthors{Li et~al.}

\title [mode = title]{Towards Generative Location Awareness for Disaster Response: A Probabilistic Cross-view Geolocalization Approach}
\tnotemark[1]

\tnotetext[1]{This is the accepted manuscript (postprint) of the following article: Li H, Deuser F, Yin W, Knoblauch S, Zhao W, Biljecki F, Xue Y, Huang W (2026). Towards generative location awareness for disaster response: A probabilistic cross-view geolocalization approach. \emph{ISPRS Journal of Photogrammetry and Remote Sensing}, 237, 130--145. The Version of Record is available at \href{https://doi.org/10.1016/j.isprsjprs.2026.03.050}{https://doi.org/10.1016/j.isprsjprs.2026.03.050}. \copyright{} 2026 Elsevier B.V. on behalf of International Society for Photogrammetry and Remote Sensing, Inc.\ (ISPRS). This manuscript version is made available under the CC BY-NC-ND 4.0 license, \href{https://creativecommons.org/licenses/by-nc-nd/4.0/}{https://creativecommons.org/licenses/by-nc-nd/4.0/}.}

\author[1]{Hao Li}[type=editor,
                        auid=000,
                        bioid=1,
                        orcid=0000-0002-6336-8772]

\cormark[1]
\ead{hao.li@nus.edu.sg}

\author[2,3]{Fabian Deuser}[type=editor,
                        auid=000,
                        bioid=1,
                        orcid=0000-0003-4511-4223]
\ead{fabian.deuser@tum.de}

\author[4]{Wenping Yin}[type=editor,
                        auid=000,
                        bioid=1,
                        orcid=0009-0002-2056-0091]
\ead{yin@cumt.edu.cn}

\author[5,6]{Steffen Knoblauch}[]
\ead{steffen.knoblauch@uni-heidelberg.de}

\author[7]{Wufan Zhao}[]
\ead{wufanzhao@hkust-gz.edu.cn}

\author[8,9]{Filip Biljecki}[orcid=0000-0002-6229-7749]
\ead{filip@nus.edu.sg}

\author[4]{Yong Xue}[]
\ead{yx9@hotmail.com}

\author[10]{Wei Huang}[]
\ead{wei_huang@tongji.edu.cn}

\affiliation[1]{organization={Department of Geography, National University of Singapore},
    city={Singapore},
    postcode={117570}, 
    country={Singapore}
    }

\affiliation[2]{organization={Institute of Distributed Intelligent Systems, University of the Bundeswehr Munich},
    city={Neubiberg},
    postcode={85579}, 
    state={Bavaria},
    country={Germany}}

\affiliation[3]{organization={Professorship of Big Geospatial Data Management, School of Engineering and Design, Technical University of Munich},
    city={Munich},
    postcode={85521}, 
    state={Bavaria},
    country={Germany}}

\affiliation[4]{organization={School of Environment and Spatial Informatics, China University of Mining and Technology},
    city={Xuzhou},
    postcode={221116}, 
    state={Jiangsu},
    country={China}}

\affiliation[5]{organization={Interdisciplinary Center for Scientific Computing, Heidelberg University},
    city={Heidelberg},
    postcode={69120}, 
    state={Baden-Württemberg},
    country={Germany}
}

\affiliation[6]{organization={Heidelberg Institute for Geoinformation Technology, Heidelberg University},
    city={Heidelberg},
    postcode={69120}, 
    state={Baden-Württemberg},
    country={Germany}
}

\affiliation[7]{organization={Urban Governance and Design Thrust, Hong Kong University of Science and Technology (Guangzhou)},
    city={Guangzhou},
    postcode={511453}, 
    state={Guangdong},
    country={China}
    }

\affiliation[8]{organization={Department of Architecture, National University of Singapore},
    city={Singapore},
    postcode={117566}, 
    country={Singapore}
    }

\affiliation[9]{organization={Department of Real Estate, National University of Singapore},
    city={Singapore},
    postcode={119245}, 
    country={Singapore}
    }
    
\affiliation[10]{organization={College of Surveying and Geo-Informatics, Tongji University},
    city={Shanghai},
    postcode={200092}, 
    country={China}}

\cortext[cor1]{Corresponding author}

\begin{abstract}
As Earth’s climate changes, it is impacting disasters and extreme weather events across the planet. Record-breaking heatwaves, drenching rainfalls, extreme wildfires, and widespread flooding during hurricanes are all becoming more frequent and more intense. Rapid and efficient response to disaster events is essential for climate resilience and sustainability. A key challenge in disaster response is to accurately and timely identify disaster locations to support decision-making and resources allocation. In this paper, we propose a Probabilistic Cross-view Geolocalization approach, called ProbGLC, exploring new pathways towards generative location awareness for rapid disaster response. Herein, we combine probabilistic and deterministic geolocalization models into a unified framework to simultaneously enhance model explainability (via uncertainty quantification) and achieve state-of-the-art geolocalization performance. Designed for rapid diaster response, the ProbGLC is able to address cross-view geolocalization across multiple disaster events as well as to offer unique features of probabilistic distribution and localizability score. To evaluate the ProbGLC, we conduct extensive experiments on two cross-view disaster datasets (i.e., MultiIAN and SAGAINDisaster), consisting diverse cross-view imagery pairs of multiple disaster types (e.g., hurricanes, wildfires, floods, to tornadoes). Preliminary results confirms the superior geolocalization accuracy (i.e., 0.86 in Acc@1km and 0.97 in Acc@25km) and model explainability (i.e., via probabilistic distributions and localizability scores) of the proposed ProbGLC approach, highlighting the great potential of leveraging generative cross-view approach to facilitate location awareness for better and faster disaster response. The data and code is publicly available at https://github.com/bobleegogogo/ProbGLC.
\end{abstract}


\maketitle

\section{Introduction}

Due to climate change and rapid urbanization, disasters are becoming more frequent and severe worldwide \citep{dietze2021flood, nohrstedt2022exploring}. Compared with the average for the last 30 years (i.e., 1994-2023), the major natural disasters in 2025 showed an 8\% increase in frequency, a 74\% increase in fatalities, a 42\% more affected population, and a 49\% increase in direct economic losses \citep{delforge2025dat}. Effective and efficient disaster response depends on rapid situational awareness, by quickly detecting affected areas, estimating damage, and understanding access constraints on the ground. A preliminary and key prerequisite for rapid situational awareness is the accurate localization of impacted areas and affected communities \citep{li2025cross}. Precise location information supports timely decision-making, effective allocation of emergency resources, and efficient recovery efforts \citep{cai2018synthesis,huang2018near,zou2023geoai,Feng_2022,2024_ijdrr_sendai}. When disaster locations are correctly identified, rescue teams can better evaluate the extent of damage, prioritize urgent needs, and deliver assistance to the most vulnerable. In this context, disaster geolocalization has emerged as a critical task.

Recently, geolocalization has progressed along two main directions: text-based and image-based methods. Text-based approaches rely on volunteered geographic information (VGI) imagery, social media posts, and other textual data sources \citep{zou2018mining,lam2023improving,yin2025llm}. Natural language processing techniques extract toponyms, link them to gazetteers, and infer locations from contextual clues when coordinates are missing \citep{hu2023survey}. These methods enable rapid identification of affected regions from large volumes of online text and provide valuable real-time information during disasters. Image-based methods commonly include ground-view image retrieval, cross-view image retrieval, and two-dimensional-three-dimensional (2D-3D) matching, designed to handle different types of image data. Earlier studies emphasized feature matching \citep{workman2015wide}, whereas recent work employs deep learning to learn shared representations between ground and overhead images \citep{lin2015learning}. Among these, cross-view geolocalization is particularly important in disaster contexts because it links crowdsourced ground images with satellite imagery to pinpoint damaged areas \citep{deuser2023sample4geo,zeng2025cross,li2025cross}. For example, street-view imagery (SVI) collected after Hurricane Ian, 2022, provided rapid situational awareness along Florida’s coast; however, moving from nadir satellite views to ground SVI perspectives remains challenging for automated geolocalization and requires cross-view techniques \citep{manzini2023harnessing}. Although advances in cross-view geolocalization have improved disaster monitoring and response, the majority of existing methods remains black boxes as deterministic approaches with limited explainability.

Explainability has become a central concern in geospatial artificial intelligence (GeoAI), as black-box models limit transparency and hinder operational deployment \citep{hsu2023explainable, li2024geoai,2024_epb_xai}. To address this issue, a variety of explainable AI (XAI) methods have been proposed. Postprocessing techniques such as saliency maps, gradient-based attribution, SHAP, and Grad-CAM aim to highlight the input regions or variables most responsible for a model’s output \citep{hohl2024recent,foroutan2025revealing}. Attention-based visualization and concept-level reasoning further provide insights into the internal decision process of AI models \citep{dehimi2024attention}. In addition, model-intrinsic approaches, including interpretable architectures, physics-aware designs, and rule-based systems, embed explainability directly into the model design \citep{hu2023updexplainer}. While these advances show strong potential for geospatial applications, their feasibility and applicability in disaster-related geolocalization remain limited: explanations are often coarse, unstable across scenarios, or difficult to interpret. This gap underscores the urgent need for domain-oriented explainability techniques that can deliver trustworthy and actionable explainability insights in disaster response.

\begin{figure}[!t]
\centering
\includegraphics[width=0.45\textwidth]{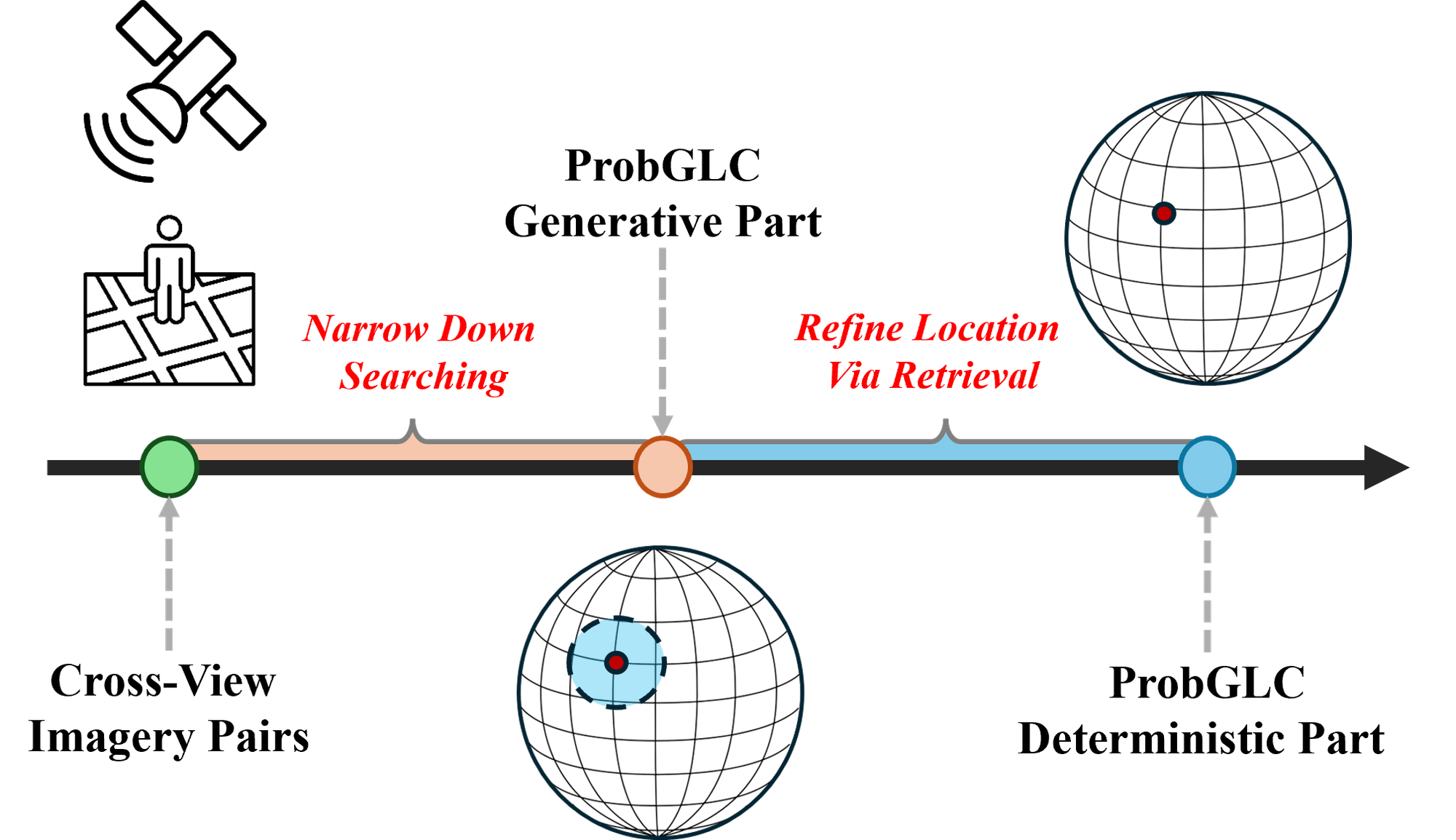}
\caption{Illustration of a simplified diagram of ProbGLC consisting of 1) generative and  2) deterministic geolocalization parts.}
\label{fig:diagram}
\end{figure}

In this context, this paper proposes a \textbf{Prob}abilistic Cross-view \textbf{G}eo\textbf{L}o\textbf{C}alization approach, namely \textbf{ProbGLC}, aiming at facilitating generative location awareness for rapid disaster response. In ProbGLC, we carefully integrate the probabilistic and deterministic geolocalization approaches into a unified framework, which simultaneously ensures model explainability (via uncertainty quantification) and achieves state-of-the-art geolocalization performance.  Tailored for rapid disaster response, ProbGLC excels the deterministic geolocalization approach by helping to narrow down the area of interest (AOI) as shown in Figure \ref{fig:diagram}, which is extremely important in disaster response scenario as every second counts when people are waiting for help. The major contributions of ProbGLC are summarized as follows:

\begin{itemize}
    \item To provide competitive geolocalization performance across multiple disaster events (e.g., hurricanes, wildfires, floods, and tornadoes).
    \item To offer unique features of density distribution and localizability scores towards a more uncertainty-aware geolocalization process.
    \item To improve the deterministic approach with a generative probabilistic prediction via Riemannian Flow Matching directly on the Sphere.
\end{itemize}

The remainder of this paper is organized as follows: Section 2 reviews related work on state-of-the-art GeoAI methods for disaster geolocalization and uncertainty quantification. Section 3 describes the proposed ProGLC method together with the generative and deterministic geolocalization approaches. Section 4 presents experimental results on two openly available disaster geolocalization datasets. Section 5 discusses key findings as well as limitations and future directions, and Section 6 concludes the paper by highlighting the main contributions.

\section{Related Work} \label{related_work}

\subsection{Climate Change and Disaster Resilience}
Climate change is intensifying the frequency and severity of extreme weather events, including hurricanes, heavy rainfall, and coastal storm surges \citep{ellis2024principles}. These phenomena are no longer isolated anomalies but increasingly recurring patterns that pose long-term risks to human settlements and infrastructure \citep{jones2012harnessing}. Rising global temperatures have led to accelerated sea-level rise and increased atmospheric moisture, both of which contribute directly to the likelihood and impact of urban flooding events caused by climate-induced disasters (e.g., Hurricanes, Heavy rainfalls, etc.) \citep{hulme1999relative}. As a result, the intersection of climate science, disaster response, and geospatial analytics has become a crucial area for research and policy intervention.

Disaster risk and resilience received insufficient emphasis in sustainable development planning, despite the obvious relationship between disasters and development \citep{hallegatte2016unbreakable, cai2018synthesis}. Disaster resilience building (e.g., prevention, preparedness, and disaster response systems) for predictable events like the major climate extreme weather conditions such as cyclones, hurricanes, wildfires, heat and cold waves \citep{li2020exploration, zhou2025rapid, shives2025multiple, russo2023increasing}, helps to protect both human and economic assets. With the growing availability of remote sensing data and advanced GeoAI methods, disaster resilience can be enhanced through near real-time monitoring, fine-grained damage mapping, and predictive modeling \citep{li2025cross}. In recent years, events like Hurricane Ian \citep{yin2025triple} and  California wildfires \citep{shives2025multiple} have highlighted the need for integrated systems that combine environmental sensing, citizen-contributed data, and AI-powered analytics to improve situational awareness and guide emergency response.

\subsection{GeoAI for Disaster Mapping and Localization}
GeoAI, the integration of geospatial data and artificial intelligence, has become a transformative force in disaster mapping and localization \citep{mai2025towards,zou2023geoai}. By combining remote sensing, computer vision, and machine learning, GeoAI enables rapid extraction of critical information from multimodal data sources. In particular, satellite-based Earth Observation (EO) and SVI provide complementary views: the former ensures wide-area coverage for macro-scale assessment, while the latter offers detailed, ground-level context \citep{biljecki2021street, zhang2024urban}. GeoAI models can fuse these views to support timely, scalable, and accurate disaster mapping across spatial resolutions and sensor modalities.

GeoAI plays a critical role in identifying the location of ground-level visual content, especially when metadata is missing or unreliable. Cross-view geolocalization, powered by deep feature matching between ground-level and overhead imagery, allows GeoAI models to infer precise locations of images or videos. This is particularly valuable in disaster scenarios where citizen-generated content or drone imagery lacks Global Positioning System (GPS) information. \cite{yin2025triple} develops two novel GeoAI methods, StaGeo and TriGeo, to improve the cross-view geolocalization accuracy of disaster-related VGI imagery based on RSI by using SVI as a bridge. \cite{vivanco2023geoclip} proposes GeoCLIP, a CLIP-inspired image-to-GPS retrieval framework that aligns images with their geographic locations using continuous spatial encoding and multi-scale feature representations for accurate worldwide geolocalization.

Moreover, recent trend in developing multi-modal and vision-language models (VLMs) sheds stimulating light on integrating large-scale remote sensing with semantic or multimodal reasoning  for disaster understanding. \cite{gupta2019creating} established a large-scale disaster mapping benchmark focusing on building damages and change detection, which largely facilitates relevant disaster model development. \cite{wang2025disasterm3} further curated a remote sensing vision-language dataset, namely DisasterM3, for global-scale disaster response, feature multi-modal visual perception and reasoning in various disaster scenes. More recent, the potential of VLMs in coordinating multi-view imagery (i.e., SVI and satellite imagery) is confirm in tackling sustainable urban development challenges \citep{wang2026cityvlm}. However, we notice a remaining research gap on examining the uncertainty and explainability of GeoAI-based disaster response models, which largely motivates the ProbGLC approach in this paper. 

\subsection{Uncertainty and Explainability of GeoAI}
Uncertainty and explainability have emerged as two central research themes in GeoAI, reflecting growing demands for reliable, transparent, and accountable decision-making in geospatial science \citep{hu2024five, li2025explainable}. GeoAI models frequently operate on heterogeneous, multi-source spatial data, introducing uncertainty from data noise, temporal dynamics, model assumptions, and output variability. At the same time, the complex nature of deep models often leads to limited explainability. Tackling both challenges is essential for enhancing model robustness, fostering trust, and enabling informed decisions in critical applications such as disaster response and climate risk management.

Uncertainty has long been recognized as a key issue in geographic information science, as both data and models are generally considered to carry uncertainty \citep{zhang2002uncertainty}. Many researchers highlight the importance of making uncertainty interpretable and actionable, where Bayesian approximation and ensemble learning are two commonly applied methods for uncertainty quantification. Bayesian techniques for uncertainty quantification include Monte Carlo dropout \citep{srivastava2014dropout}, Markov chain Monte Carlo \citep{kupinski2003ideal}, and Variational inference \citep{blei2017variational}. Common ensemble learning methods include deep ensemble, deep ensemble Bayesian/Bayesian deep ensemble, and uncertainty in Dirichlet deep networks \citep{abdar2021review}. Moving forward, integrating explainability and uncertainty handling will be crucial for advancing the scientific and practical impact of GeoAI.

However, while uncertainty quantification is crucial, the explainability of GeoAI models remains a challenge, especially in spatial applications like environmental monitoring and disaster management \citep{li2025explainable}. Traditional “black-box” models often fail to provide intuitive explanations of their decision-making processes, making it difficult to trust their outputs. To address this, research has increasingly focused on post-hoc explainability methods, such as local interpretable model-agnostic explanations \citep{ribeiro2016should}, Shapley additive explanations (SHAP) \citep{lundberg2017unified}, and class activation mapping (CAM) \citep{zhou2016learning}, which provide insights into model behavior by attributing output decisions to specific input features.  However, these methods primarily focus on interpreting outputs rather than understanding the internal mechanisms of geospatial models. Therefore, there is growing emphasis on developing intrinsically interpretable models that integrate spatial knowledge, physical principles, and domain expertise, while also improving uncertainty and explainability methods in GeoAI models.

\section{Methodology} \label{method}
\begin{figure*}[!t]
\centering
\includegraphics[width=\textwidth]{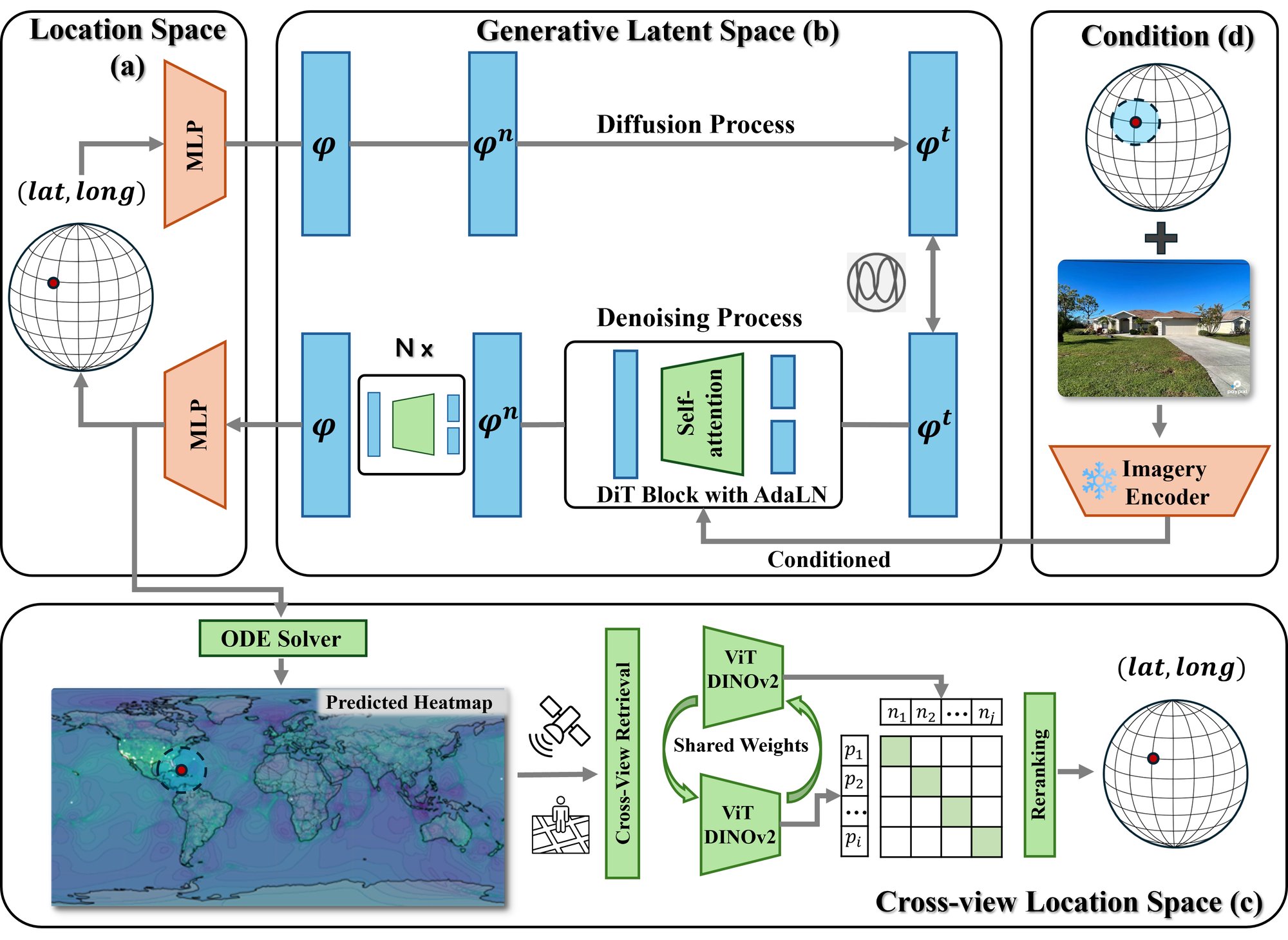}
\caption{Overview of the \textbf{Prob}abilistic Cross-view \textbf{G}eo\textbf{L}o\textbf{C}alization (\textbf{ProbGLC}) approach. The ProbGLC consists of mainly four part: (1) Location Space with geographical coordinates (lat and long) of disaster-related imagery; (2) Generative Latent Space where the generative model is trained to learn a latent space for probabilistic geolocalization tasks; (3) Cross-view Location Space with the deterministic retrieval approach is trained to refine the location predictions; (4) Condition of disaster-related imagery embeddings and model hyperparameter. Herein, $\varphi$ refers to diffusion model parameters, $p_i$ and $n_i$ are cross-view imagery embeddings.}
\label{fig:task_statement}
\end{figure*}

\subsection{Notation and Mission Statement}
Given a set of disaster-related VGI imagery $\{ L_c \}$ and satellite imagery $\{ L_s \}$, we aim to predict the most likely location $x_0$ (longitude and latitude) where the imagery pair was taken. Unlike the pure deterministic approach, we model the conditional probabilistic distribution $P(x_n \mid \mathbf{c})$ as a generative process, where $x_n$ can be any point on Earth, modeled as the unit sphere $\mathbb{S}^2$ in $\mathbb{R}^3$, and $\mathbf{c}$ is the imagery embedding of the VGI imagery used as a query input. Herein, the random noise is denoted as $\varepsilon$ and the noisy coordinates as $x_t$ for a timestep $t$. Moreover, $\mathbb{\varphi}$ refers to a set of trainable parameters in the neural network (Figure \ref{fig:task_statement}). 

Based on the probabilistic distribution $P(x_n \mid \mathbf{c})$,  we then run the cross-view imagery retrieval as a deterministic process, but only within a refined area of interest radiused with $\mathbf{r}$. This deterministic geolocalization process learns a cross-view embedding space $\R_{CV}$ so that the exact geographical location of VGI imagery $\{ L_c \}$ can be retrieved based on its similarities to geo-tagged satellite imagery $\{ L_s \}$ in the learned embedding space. Figure \ref{fig:task_statement} shows how we achieve the generative location awareness with a probabilistic geolocalization approach, leveraging cross-view imagery and state-of-the-art GeoAI models to facilitate the critical disaster response scenario. In the rest of the section, we will elaborate on the detailed methodology and design specifics.

\subsection{Generative Probabilistic Geolocalization}
In the disaster response context, modeling geolocalization as a generative process has mainly two benefits: 1) to explicate and quantify uncertainty of visual geolocalization which mostly remains a black-box and deterministic process, 2) to offer a generalizable approach of addressing spatial ambiguity across multiple disaster types. In the rest of this section, we will elaborate on how \textbf{ProbGLC} models the probabilistic geolocalization process using a Riemannian Flow Matching diffusion approach, which is inspired by \citet{dufour2024around}.

\begin{figure}[!t]
\centering
\includegraphics[width=\columnwidth]{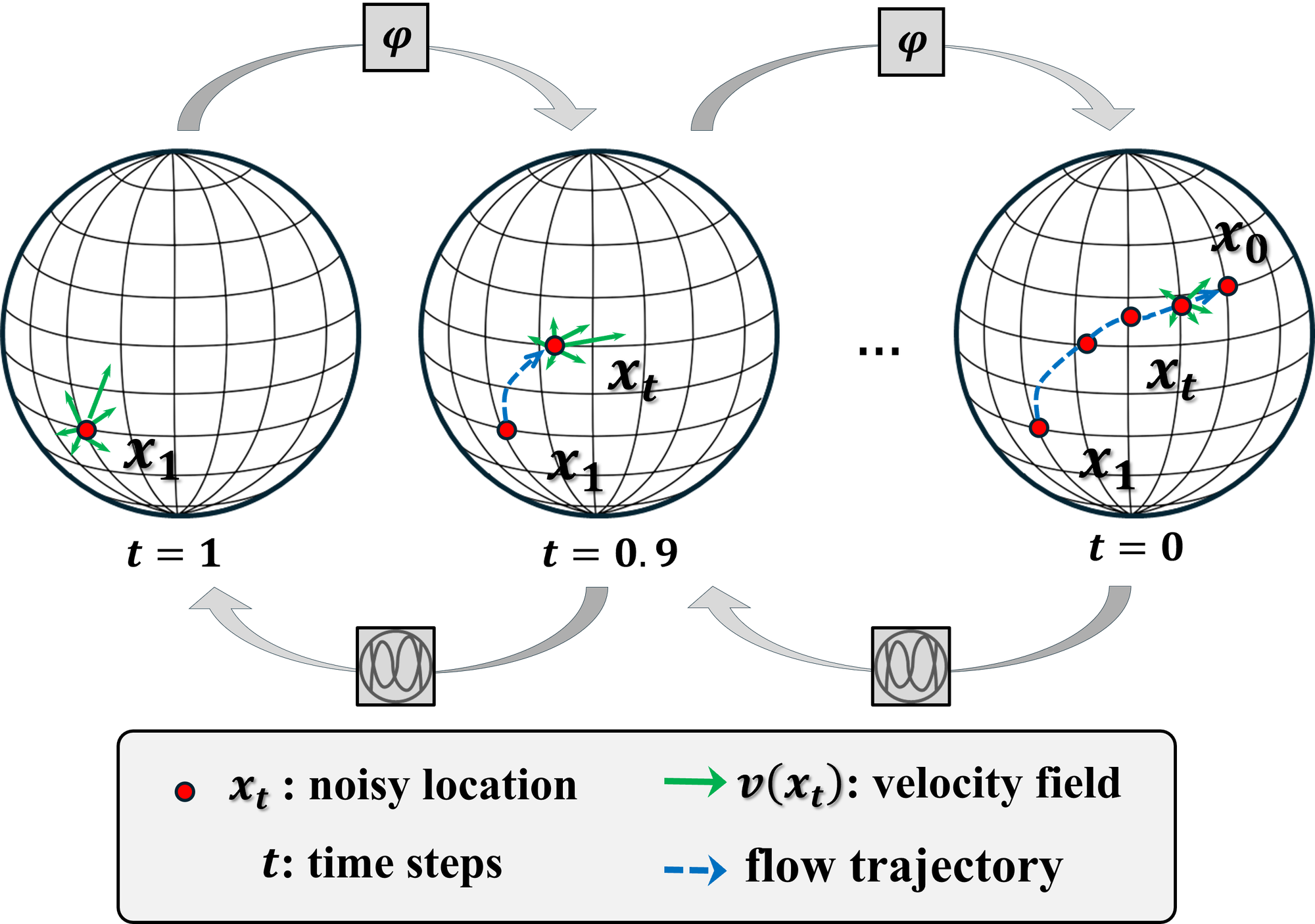}
\caption{RFM on the sphere. Visualization of the generative flow process where noisy locations evolve along a learned velocity field toward their denoised positions on the spherical manifold.}
\label{fig:RFM}
\end{figure}

\subsubsection{Riemannian Flow Matching on the Sphere}

The previous steps model the classic imagery geolocalization task as a generative process that can be trained with state-of-the-art diffusion training pipelines.
Figure \ref{fig:task_statement} shows the overview workflow in the ProbGLC with part (b) refers to a generative diffusion model for imagery geolocalization. Specifically, we employed a Riemannian Flow Matching (RFM) approach from \citet{dufour2024around} to extend the classic geographic diffusion process (see Appendix \ref{Appen_A}) allowing training and inferencing directly on the Sphere as shown in Figure \ref{fig:RFM}.

\textbf{Flow Matching in the Euclidean space}:  Classic diffusion process is built on Denoising Diffusion Probabilistic Models (DDPM). DDPM is useful but can be expensive and difficult to train as the convergence is heavily dependent on the noise schedule function $\beta(t)$. Flow matching offers an elegant solution to this problem \citep{lipmanflow}.

As a context, in diffusion models, the forward process adds noise, and the backward process removes noise. Both the forward and backward processes are regarded as Stochastic Differential Equation (SDEs), though the forward process is integrable in closed-form, so it can be done at no computational cost. However, the backward denoising process is not integrable in closed-form, so it must be integrated step-by-step using standard SDE solvers, which makes it computationally expensive. In flow-based diffusion models, the forward process will be a deterministic flow along a time-dependent vector field $v(\mathbf{x}_t)$, and the backward process is redefined also a deterministic flow along the very same vector field, but backwards. To this end, both processes become Ordinary Differential Equation (ODE). This means if the vector field is well-behaved, the ODE will also be well-behaved, making the training easier and faster. 

Given the time-dependent vector field $v(\mathbf{x}_t)$, it can be calculated as the partial derivative of the noisy location (namely Equation \ref{eq1}) over the timestep $t$:

\begin{equation} \label{eq9}
v(\mathbf{x}_t) = \frac{d \mathbf{x}_t}{d t} = \dot{\frac{d\beta(t)}{dt}(\varepsilon - \mathbf{x}_0)},
\end{equation}

The training target of flow matching is then to minimize the prediction error of $\mathbb{\varphi}(\mathbf{x}_t \mid c)$ to the vector field similarly as in Equation \ref{eq6}:

\begin{equation} \label{eq10}
\mathcal{L}_{FM} = \mathbb{E}_{\mathbf{x}_0,  \mathbf{c}, \epsilon, t} \left[ || \mathbb{\varphi}(\mathbf{x}_t \mid \mathbf{c}) - v(\mathbf{x}_t )||^2 \right],
\end{equation}

\textbf{Inference Process}: For inference, one can then use ODE solver initialized at a random location $\mathbf{x}_1 = \epsilon$, integrating backward from $t = 1$ to $t = 0$ using the trained parameters $\mathbb{\varphi}(\mathbf{x}_t \mid c)$:

\begin{equation} \label{eq11}
    \mathbf{x}_{t - \Delta t} = \mathbf{x}_t - \mathbb{\varphi}(\mathbf{x}_t \mid c)\, \Delta t
\end{equation}

But noticed this is still in the Euclidean space, and one will have to project the final location $\mathbf{x}_0$ back to the Spherical coordinates of $\mathbb{S}^2$.

\textbf{Extend to Riemannian Flow Matching on the Sphere}: To constrain the noisy location $\mathbf{x}_t$ always on the Sphere $\mathbb{S}^2$ without of projections, the trick is to extend the aforementioned flow matching to Riemannian manifolds \citep{chenflow} which ensure both the location variable $\mathbf{x}_t$ and the noise $\epsilon$ remain on $\mathbb{S}^2$. The former one is easy as the natural format of a location is in spherical coordinates, and the latter one can be achieved by manipulating Equation \ref{eq1} as follows:

\begin{equation} \label{eq12}
    \mathbf{x}_t = \exp_{\mathbf{x}_{t-1}}\!\big( \beta(t) \, \log_{\mathbf{x}_{t-1}}(\epsilon) \big),
\end{equation}

Where $\log_{\mathbf{x}_{t-1}}$ is the logarithmic map, mapping a point on $\mathbb{S}^2$ to the tangent space at a previous timestep $\mathbf{x}_{t-1}$, and $\exp_{\mathbf{x}_{t-1}}$ refers to the exponential map back to the Riemannian manifold. For more details about the projection, see Appendix \ref{Appen_b}.

This representation will allow us to keep the noisy coordinates always along the geodesic, and let us rewrite the vector field on the tangent space of $\mathbf{x}_t$ as follows:

\begin{equation} \label{eq13}
    v(\mathbf{x}_t) = \frac{d \mathbf{x}_t}{d t} = \dot{\frac{d\beta(t)}{dt}\log_{x_t}(x_0)},
\end{equation}

Where $\log_{x_t}(x_0)$ refers to the  tangent vector at $\mathbf{x}_t$ pointing along the geodesic from $\mathbf{x}_0$ to $\epsilon$ (as shown in Figure \ref{fig:RFM}). Similarly to Equation \ref{eq8}, we can predict the velocity field by minimizing the prediction error constrained by the Riemannian manifold:

\begin{equation} \label{eq14}
    \mathcal{L}_{RFM} = \mathbb{E}_{\mathbf{x}_0,  \mathbf{c}, \epsilon, t} \left[ || \mathbb{\varphi}(\mathbf{x}_t \mid c) - v(\mathbf{x}_t) ||^2_{\mathbf{x}_t} \right],
\end{equation}

Where the only difference compared to Equation \ref{eq8} is the $\lVert \cdot \rVert_{x_t}$ denoting the norm induced by the Riemannian metric on the tangent space at $\mathbf{x}_t$.

\textbf{Probabilistic Inference on the Sphere}: To predict with RFM, once the $\mathbb{\varphi}(\mathbf{x}_t \mid c)$ is trained, we can use it to compute the geolocalization probability of any location on the Sphere as $P(x_n \mid \mathbf{c})$ conditional on a query imagery embedding $c$. This enables us to explicate and quantify the uncertainty of visual geolocalization, contrasting with classic approaches, which mostly remain deterministic. The probability distribution is calculated in the following way:

\begin{equation} \label{eq15}
    \log P(x_n \mid \mathbf{c}) = \log P_{\epsilon}(x(1) \mid \mathbf{c}) - f(1),
\end{equation}

Where $P_{\varepsilon}$ is the known distribution of the pure noise $\epsilon$. And $f(t)$ is a function that accumulates the negative divergence of the velocity field along the trajectory $x_t$ using the following ODEs:

\begin{equation} \label{eq16}
\frac{d}{dt}
\begin{bmatrix}
x_t \\
f(t)
\end{bmatrix}
=
\begin{bmatrix}
\mathbb{\varphi}(\mathbf{x}(t) \mid c) \\
- \operatorname{div} \mathbb{\varphi}(\mathbf{x}_t \mid c)
\end{bmatrix},
\quad \text{with} \quad
\begin{bmatrix}
\mathbf{x}_0 \\
f(0)
\end{bmatrix}
=
\begin{bmatrix}
\mathbf{x}_n \\
0
\end{bmatrix}.
\end{equation}

For a detailed explanation, please refer to the original implementation in TorchDiffEq\footnote{https://github.com/rtqichen/torchdiffeq}. This probabilistic inference directly on the sphere finally allows us to quantify the uncertainty and calculate the "localizability" score of the generative geolocalization process. Following the idea from \citet{dufour2024around}, we define the "localizability" score as a measure of spatial ambiguity in geolocalization using the following equation: 

\begin{equation} \label{eq17}
    \mathrm{Localizability}(\mathbf{c}) = \int_{\mathbb{S}^2} p(\mathbf{x}_n \mid \mathbf{c}) \log_{2} P(\mathbf{x}_n \mid \mathbf{c}) \, d\mathbf{x}_n, 
\end{equation}

Where the localizability of an image $c$ is the negative entropy of the predicted distribution. In practice, we estimate this integral with Monte-Carlo sampling \citep{metropolis1949monte} using 10,000 samples.

\subsection{Cross-view Deterministic Geolocalization}
In this work, \textbf{ProbGLC} takes the generative probabilistic geolocalization as a first step, then combines with a deterministic cross-view geolocalization approach only with a certain radius thresholding by $\mathbf{r}$. Herein, the consideration is mainly twofold: 1) though the RFM approach offers a unique feature of uncertainty quantification and spatial ambiguity measurement, the deterministic approach still provide a more fine-grained geolocalization accuracy comparing to the generative approach with the cost of additional computational effort; 2) taking advantage of the probabilistic distribution generated from the RFM approach, we can improve the deterministic approach by narrowing down the area for imagery retrieval, which is extremely important in disaster response scenario as every second counts when people are waiting for help. The rest of this section will elaborate on the deterministic cross-view geolocalization approach.

\subsubsection{Siamese Cross-view Image Encoder}
Herein, we use a Siamese Vision Transformer with DINOv2 (i.e., ViT DINOv2) encoder to learn a cross-view embedding space $\R_{CV}$ so that the exact geographical location of VGI imagery $\{ L_c \}$ can be retrieved based on its similarities to geo-tagged satellite imagery $\{ L_s \}$. Specifically, ViT DINOv2 treats each image as a sequence of fixed-size patches and captures global dependencies through a multi-layer self-attention \citep{dosovitskiy2020image}. As a state-of-the-art vision model, ViT DINOv2 provides a strong and generalizable backbone for the deterministic cross-view geolocalization approach, where a range of previous studies have confirmed its semantic representation capability, robustness to large viewpoint changes, and cross-domain generalizability \citep{oquab2023dinov2,yin2025triple}.

Moreover, we adopt a Siamese design for learning the cross-view embedding space $\R_{CV}$. As shown in the Figure. \ref{fig:task_statement}, the same backbone encoder (i.e., ViT DINOv2) is used, but with independent weights for each data modality. This design addresses the issue of different image sizes across modalities, where RSI and VGI images can share a similar encoder. Next, the weight-sharing strategy across other modalities further guarantees that both inputs are embedded into a common feature space under the same semantic criteria \citep{bromley1993signature}, which is crucial for accurate cross-view imagery matching and retrieving. Finally, this Siamese imagery encoder, denoted as $f()$, is trained on the imagery pair of VGI and RSI with a contrastive learning objective as follows:

Given a context vector $c$, the positive sample is drawn from a conditional distribution $p(x|c)$, where $(N-1)$ negative samples are drawn from the same distribution $p(x)$ but without condition. In this context, the probability of correctly selecting the positive samples can be formulated as follows:

\begin{equation} \label{eq18}
    P(C=\texttt{pos} \vert X, \mathbf{c}) = \frac{f(\mathbf{x}_\texttt{pos}, \mathbf{c})}{ f(\mathbf{x}_\texttt{pos}, \mathbf{c}) + \sum_{j=1}^{N-1} f(\mathbf{x}_j, \mathbf{c}) }
\end{equation}

Here, N is the total number of samples in a batch, and $f(\mathbf{x}, \mathbf{c}) \propto \frac{P(\mathbf{x}\vert\mathbf{c})}{P(\mathbf{x})}$ is the similarity or scoring function between two samples. 

Then, the contrastive InfoNCE loss can be calculated as follows: 

\begin{equation} \label{eq19}
    \mathcal{L}_\text{InfoNCE} 
    = - \mathbb{E} \left[\log p(C=\texttt{pos} \vert X, \mathbf{c}) \right]
\end{equation}

Where the objective of $\mathcal{L}_\text{InfoNCE} $ is to optimize the negative log probability of selecting the positive samples. 

In this work, we use the InforNCE as our contrastive learning loss in both the pre-training and fine-tuning stages for deterministic cross-view geolocalization. During the fine-tuning, we take the model weights pre-trained on CVUSA data given its relatively large size and geographical closeness \citep{workman2015wide}, then fine-tune the model on the cross-view imagery pairs collected from two disaster mapping datasets.

\subsubsection{Self-Supervised Feature Learning with DINOv2}

The cross-view image encoder used in this work is pretrained with DINOv2, a state-of-the-art self-supervised learning method based on the teacher-student architecture \citep{dosovitskiy2020image}. Herein, DINOv2 enables the model to learn semantically rich and domain-invariant features by enforcing consistency between representations of diverse augmented views of the same image, without additional supervision signals. This is achieved by minimizing a cross-view consistency loss between the teacher and student outputs \citep{oquab2023dinov2}. The self-supervised loss function for DINO is defined as follows:

\begin{equation} \label{eq20}
\mathcal{L}_{DINO} = - \sum_{i=1}^{N} P^{(t)}(x_i) \cdot \log P^{(s)}(x_i)
\end{equation}

Where $P^{(t)}(x_i)$ and $P^{(s)}(x_i)$ are the teacher and student output distributions for the $i$-th feature, respectively, and $N$ is the number of feature dimensions. This training objective optimizes the student network to align its outputs with those of the teacher across varying augmentations, resulting in stable and semantically-consistent representations of input data from diverse view angles.

In this context, DINOv2 enhances the ViT backbone’s generalization ability across heterogeneous modalities from ground-view to head-view, which often differ in their resolution, granularity and scene complexity. Compared to CNN-based models, ViT pretrained with DINOv2 tends to reserve long-range dependencies and semantic coherence, making it particularly effective for aligning multi-view features \citep{oquab2023dinov2}. In this context, we use the same encoder architecture to encode both VGI and RSI imagery to ensure a consistent feature extraction and facilitate efficient joint training, ultimately improving geolocalization accuracy and reliability in complex disaster response scenarios.

\begin{figure}[!t]
\centering
\centering
\includegraphics[width=\columnwidth]{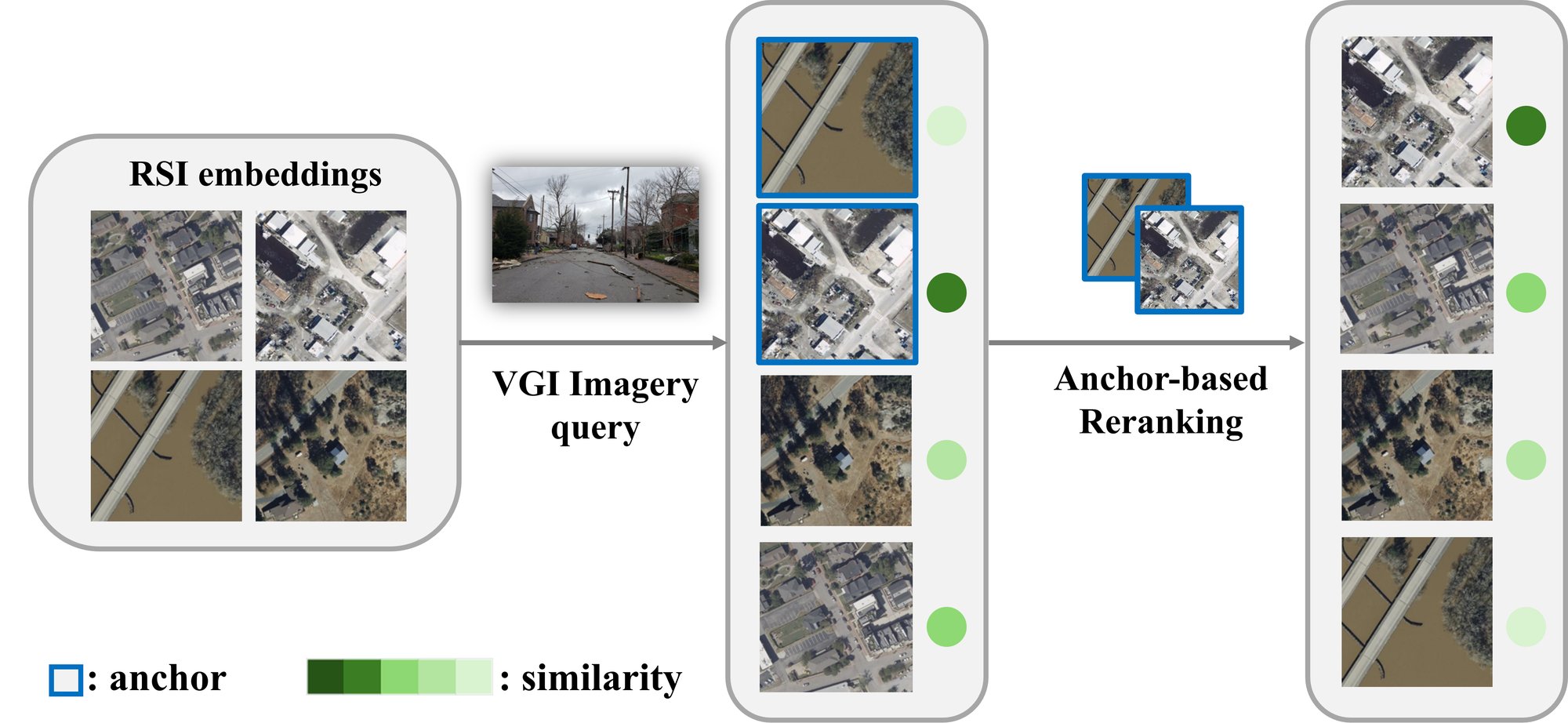}
\caption{Anchor-based reranking for cross-view geolocalization. Retrieved RSI candidates are refined using anchor samples to improve similarity consistency with the ground-view query.}
\label{fig:reranking}
\end{figure}

Moreover, to further improve the cross-view geolocalization accuracy, we add an anchor-based reranking strategy \citep{deuser2024optimizing} as a post-processing step after the initial retrieval (See Appendix \ref{Appen_c}). As a result, the reranking process enforces local neighborhood consistency and mitigates potential mismatches from the initial retrieval. For more details ablation, one can refer to \cite{deuser2024optimizing}.

\section{Experiment} \label{experiment}

\begin{figure*}[!t]
\centering
\centering
\includegraphics[width=\textwidth]{figures/Dataset_multi.jpg}
\caption{Spatial distribution and examples of the SAGINDisaster dataset. (a) Distribution of samples for different disaster types. (b) Example images of each disaster type, with border colors matching the categories in (a).}
\label{fig:Dataset_multi}
\end{figure*}

\begin{figure}[!t]
\centering
\centering
\includegraphics[width=\columnwidth]{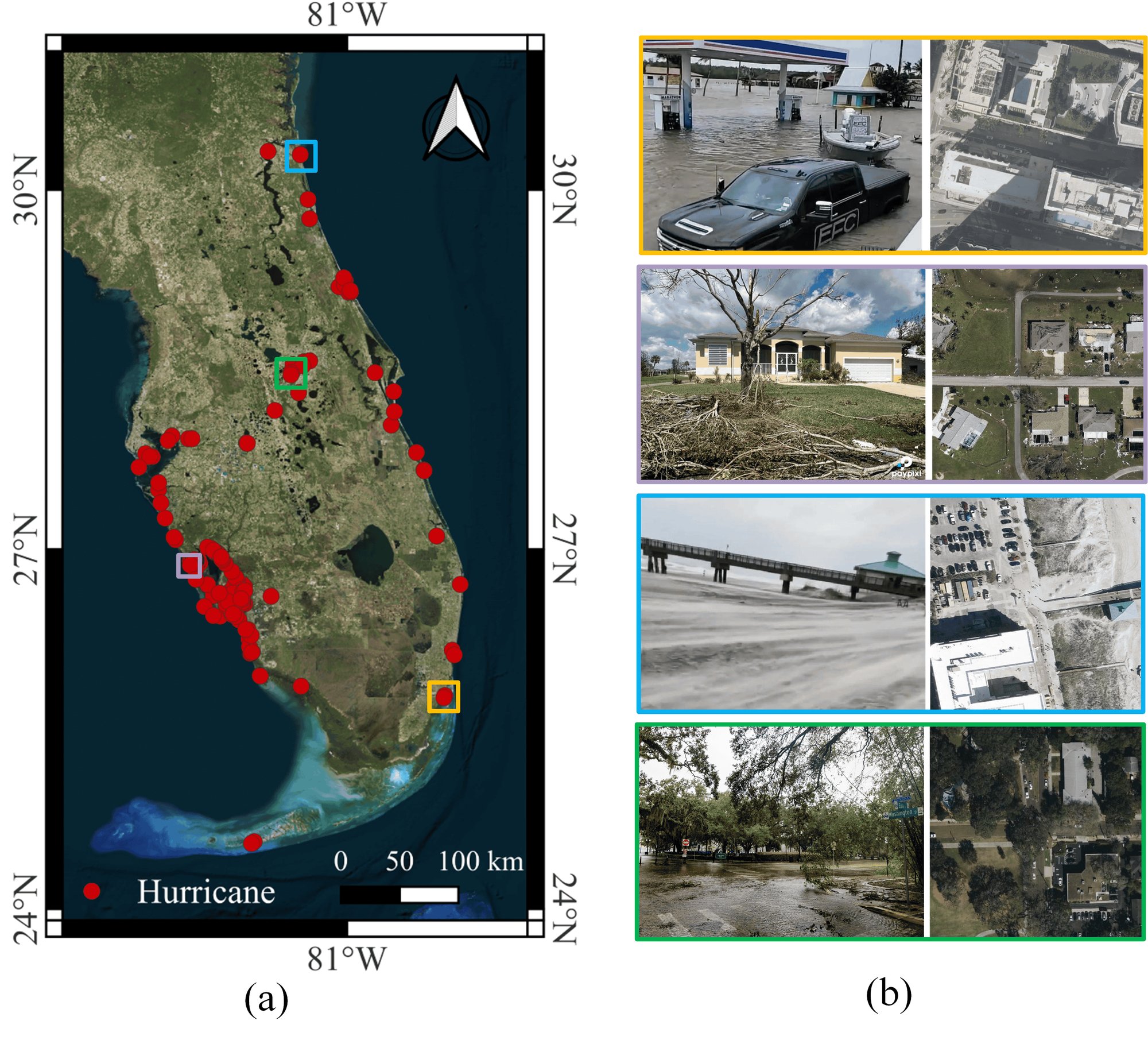}
\caption{Spatial distribution and examples of the MultiIAN dataset. (a) Distribution of samples for different disaster types. (b) Example scenes of VGI and Satellite image pairs, where colors refer to locations in (a).}
\label{fig:Dataset_ian}
\end{figure}

\subsection{Dataset Overview} \label{data_overview}
In this work, we evaluate the proposed approach against two openly available disaster geolocalization datasets, namely MultiIAN and SAGINDisaster datasets, with a different geographical focus and disaster types as shown in Figure \ref{fig:Dataset_multi} and \ref{fig:Dataset_ian}. 

For the MultiIAN dataset \citep{yin2025triple}, it focuses on southeastern United States and western Cuba following the catastrophic Hurricane Ian, a Category 5 Atlantic hurricane occurring from late September to early October 2022. It reached peak intensity on September 28 near southwestern Florida, and became the third costliest weather disaster globally \citep{Wikipedia2022Ian}. As a result, MultiIAN includes 5,706 imagery pairs containing both VGI and RSI imagery.

As for SAGINDisaster dataset \citep{Yin2025SAGINGeo}, multiple representative disaster types are selected across the continental United States including the Hawaiian Islands to diversify the potential cross-view disaster geolocalization scenarios. As a context, we take an advantage of the emergency response imagery (over 50 disaster events) released by the National Oceanic and Atmospheric Administration (NOAA) Remote Sensing Division\footnote{https://storms.ngs.noaa.gov/}, ranging from hurricanes, wildfires, floods, to tornadoes. Specifically, these disasters include the California Fire (2025) and Maui Fire (2023), the aforementioned Hurrican IAN (2025), the Nashville Tornadoes (2020) and Illinois Tornadoes (2015), and the Louisiana Flooding (2016) and Midwest U.S. Flooding (2015). In this context, SAGINDisaster results in a total of 2,080 cross-view imagery pairs (as shown in Figure \ref{fig:Dataset_multi}), featuring the diversity and challenge of cross-view geolocalization in real-world disaster response scenarios.

\textbf{VGI:} VGI imagery was collected from multiple online platforms where people voluntarily shared images and posts during and after the disasters. For Hurricane Ian, the Paypixl platform provided a large number of geo-tagged photos. In addition, we included images from Twitter and Flickr. Though Paypixl and Flickr already contain location metadata, the Twitter samples required manual annotation to determine their geographic position. These crowdsourced data reflect real on-site observations from affected communities, which are key for effective and rapid disaster responses.

\textbf{RSI:} High-resolution satellite images were sourced from the U.S. National Oceanic and Atmospheric Administration (NOAA). These images were captured during the disaster periods, especially Hurricane Ian, and processed into accessible mosaics and tiles. With a ground sampling distance of 15–30 cm per pixel, the satellite data offer a detailed overview of damaged areas and provide a critical large-scale perspective to complement ground-level information. Serving as the most important reference imagery in cross-view geolocalization, RSI plays a crucial role in aligning and anchoring information from a head-view perspective.

\subsection{Evaluation Metrics and Implementations}
The main idea of ProbGLC is to combine the probabilistic and deterministic geolocalization approaches to simultaneously achieve uncertainty quantification and performance boosting. In this context, the experiment aims to validate a two-level improvement against a range of baseline models and variations. First, the improvement over pre-trained generative approaches, for example in \citet{dufour2024around} without fine-tuning for disaster-related VGI imagery; second, the improvement over either solely fine-tuned generative approaches or classic imagery retrieval approaches. 

For generative baselines, we consider three types of pre-trained visual geolocalization models, namely 1) Diffusion: the geographic diffusion approach in the Euclidean space (Appendix \ref{Appen_A}), 2) Flow Matching in Euclidean space with time-dependent vector field, and 3) Riemannian Flow Marching (RFM) directly on the Sphere $\mathbb{S}^2$ as suggested in \citet{dufour2024around}. Herein, their pre-training weights are obtained from three state-of-the-art global visual geolocalization datasets (i.e., YFCC \citep{thomee2016yfcc100m}, iNaturalist \citep{van2021benchmarking}, OSV-5M \citep{astruc2024openstreetview}), respectively. 

Moreover, to investigate the impact of combining probabilistic and deterministic geolocalization approaches, we carefully design a set of comparisons between the ProbGLC approach with either solely probabilistic or deterministic geolocalization approaches, especially with fine-tuning on the disaster-related cross-view datasets. We examine the pattern of our performance gain under different thresholds $\mathbf{r}$, which is considered as a key hyperparameter of the ProbGLC approach to balance accuracy and explainability from either  deterministic or probabilistic approaches. For deterministic cross-view geolocalization approaches, we include the state-of-the-art models as baselines as well, namely SAIG-D \citep{zhu2023simple} and TransGeo (ViT-based) \citep{zhu2022transgeo}.

To evaluate the model performance, since the ProbGLC approach is one the very first to tackle the geolocalization as a generative process rather than classic imagery retrieval, we have to use a novel metrice called Acc@K rather than the classic Recall@K as he main performance metric. Herein, Acc@K measures the proportion of correct matches (between predicted locations and exact locations) within
a certain geographical radius. In this study, Acc@1km, Acc@25km, Acc@50km, and Acc@200km are used to measure the performance of both probabilistic and deterministic approaches. In addition, we also report the median and mean distances over the test set of our disaster cross-view datasets. All evaluations are performed using both 8:2 and 7:3 train-test splits for two datasets, respectively.

A unique metric of the ProbGLC approach is the localizability score featuring a quantitative measure of how difficult it is to geolocalize the disaster-related VGI imagery with the generative approach using the Riemannian Flow Marching. This localizability score in combination with the density maps (see in Figure \ref{fig:task_statement}) offers novel insights towards generative location awareness for disaster response.

Regarding the training and implementation, for ProbGLC variations, we use a learning rate of $8 \times 10^{-4}$ and a step size ($\Delta t$) of 16 with the LAMB optimizer \citep{you2019large}. This choice is motivated by the large batch size of 512. The learning rate is scheduled using cosine decay down to zero. All experiments are implemented in PyTorch \citep{paszke2019pytorch} and conducted on a DGX-2 system equipped with with 16 GPUs (i.e., Nvidia V100), each with 32 GB of memory. For the ProbGLC architecture, we adopt the same network design as \cite{dufour2024around}, consisting of an MLP with 8 layers and a hidden dimension of 256. The trained retrieval model follows the Sample4Geo architecture \citep{deuser2023sample4geo} and employs a ConvNeXt-Base backbone \citep{liu2022convnet}. It is trained with a learning rate of $1 \times 10^{-3}$, a cosine decay learning rate scheduler, and a batch size of 128.

\subsection{Baseline Geolocalization Results} \label{baseline result}

\textbf{Deterministic baselines:} Table \ref{tab:detbaseline} presents the baseline performances of three state-of-the-art cross-view geolocalization approaches (i.e., TransGeo, SAIG-D, and Sample4Geo) across the two diaster datasets (i,e, SAGINDisaster, MultiIAN) using the generative evaluation metrics (Acc@K). Specifically, we take the Recall@1 match from the classic imagery-retrieval models as predicted locations then measure the distance to actual locations of VGI imagery. 

As a baseline for deterministic approaches, ProbGLC demonstrates the strongest zero-shot performance across both datasets, showing notably higher retrieval accuracy and lower localization errors.For SAGINDisaster dataset (20\% test), it achieves Acc@50km of 0.632 and Acc@200km of 0.659, while reducing the mean distance to 903.68 km and the median distance to only 3.40 km. Similarly for MultiIAN (20\% test), Sample4Geo also excels with Acc@25km reaching 0.963 and Acc@200km reaching 0.990, alongside very low mean and median distances of 7.26 km and 2.40 km. In contrast, TransGeo and SAIG-D yield lower accuracies, such as Acc@50km of 0.339 and 0.387 on SAGINDisaster, with much higher mean distances exceeding 1500 km.
However, a key finding herein is that all baseline approaches performs poorly for Acc@1km with an average accuracy level below 0.150 despite of datasets, which highlight the pressing need for closing this "last kilometer" gap in diaster geolocalization. 

One can easily notice the fact that two datasets (i,e, SAGINDisaster, MultiIAN) are not equally difficult. As shown in Figure \ref{fig:Dataset_multi} and \ref{fig:Dataset_ian}, SAGINDisaster includes multiple disaster types and events across the continental United States, therefore it is considered as more challenging than MultiIAN when it comes to geolocalization accuracy. This fact is also confirmed in Table \ref{tab:detbaseline}, resulting in much lower Mean and Median Dist for MultiIAN that all falls in 50 km range. However, as a nature of deterministic approach, the lack of explainability and uncertainty quantification during cross-view geolocalization often makes performance boosting an exhaustive and time-demanding task.

\begin{table*}[ht] 
\centering
\caption{Zero-shot retrieval results on the Multidisaster and IANdisaster datasets for 20\% and 30\% splits.}
\label{tab:detbaseline}
\begin{adjustbox}{width=1\linewidth}
\begin{tabular}{llccccccc}
\toprule
Dataset & Model & Split & Acc@1km & Acc@25km & Acc@50km & Acc@200km & Mean Dist (km) & Median Dist (km) \\
\midrule
\multirow{6}{*}{SAGINDisaster} 
& TransGeo   & 20\% & 0.062 & 0.219 & 0.339 & 0.440 & 1517.75 & 789.76 \\
& TransGeo   & 30\% & 0.045 & 0.191 & 0.300 & 0.426 & 1404.18 & 753.69 \\
& SAIG-D     & 20\% & 0.053 & 0.226 & 0.387 & 0.423 & 1746.45 & 810.78 \\
& SAIG-D     & 30\% & 0.040 & 0.186 & 0.285 & 0.413 & 1709.95 & 886.49 \\
& Sample4Geo & 20\% & \textbf{0.137} & \textbf{0.623} & \textbf{0.632} & \textbf{0.659} & \textbf{903.68} & \textbf{3.40} \\
& Sample4Geo & 30\% & 0.125 & 0.556 & 0.566 & 0.599 & 1302.82 & 3.79 \\
\midrule
\multirow{6}{*}{MultiIAN} 
& TransGeo   & 20\% & \textbf{0.132} & 0.583 & 0.735 & 0.946 & 39.91 & 3.34 \\
& TransGeo   & 30\% & 0.130 & 0.606 & 0.762 & 0.947 & 38.07 & 3.39 \\
& SAIG-D     & 20\% & 0.096 & 0.475 & 0.715 & 0.960 & 40.26 & 27.54 \\
& SAIG-D     & 30\% & 0.095 & 0.494 & 0.710 & 0.961 & 39.76 & 26.87 \\
& Sample4Geo & 20\% & 0.130 & \textbf{0.963} & \textbf{0.967} & \textbf{0.990} & \textbf{7.26} & \textbf{2.40} \\
& Sample4Geo & 30\% & 0.117 & 0.954 & 0.961 & 0.988 & 8.38 & 2.41 \\
\bottomrule
\end{tabular}
\end{adjustbox}
\end{table*}

\begin{table*}[h]
\centering
\caption{Zero-shot performance of RFM-70 and RFM-80 models on the SAGINDisaster dataset across different pre-training datasets.}
\label{tab:genbaseline_multi}
\begin{adjustbox}{width=1\linewidth}
\begin{tabular}{l l cccccc}
\hline
Pre-training & Method & Acc@1km & Acc@25km & Acc@50km & Acc@200km & MeanDist (km) & MedianDist (km) \\
\hline
\multirow{6}{*}{iNaturalist}
 & Diffusion-70 & 0.000 & 0.008 & 0.035 & \textbf{0.269} & \textbf{1763.69} & \textbf{534.15} \\
 & Diffusion-80 & 0.000 & 0.005 & 0.024 & 0.231 & 1777.19 & 758.33 \\
 & FlowR3-70    & 0.000 & \textbf{0.014} & \textbf{0.040} & 0.212 & 2493.61 & 1221.62 \\
 & FlowR3-80    & 0.000 & 0.007 & 0.029 & 0.209 & 2305.87 & 1317.18 \\
 & RFM-70       & 0.000 & 0.005 & 0.016 & 0.205 & 1800.86 & 770.04 \\
 & RFM-80       & 0.000 & 0.007 & 0.019 & 0.202 & 2030.33 & 818.69 \\
\hline
\multirow{6}{*}{YFCC}
 & Diffusion-70 & 0.000 & 0.038 & 0.080 & 0.516 & 1136.12 & 195.25 \\
 & Diffusion-80 & 0.000 & 0.034 & 0.084 & 0.524 &  926.89 & 195.12 \\
 & FlowR3-70    & 0.002 & 0.058 & 0.119 & 0.429 & 1065.15 & 221.77 \\
 & FlowR3-80    & 0.002 & 0.053 & 0.094 & 0.394 & 1060.01 & 226.22 \\
 & RFM-70       & 0.006 & 0.054 & 0.096 & 0.537 & 1078.54 & 187.37 \\
 & RFM-80       & \textbf{0.014} & \textbf{0.055} & \textbf{0.101} & \textbf{0.558} & \textbf{941.71} & \textbf{165.56} \\
\hline
\multirow{6}{*}{OSV-5M}
 & Diffusion-70 & 0.000 & 0.021 & 0.099 & 0.752 &  515.18 & 106.70 \\
 & Diffusion-80 & 0.000 & 0.026 & 0.099 & 0.762 &  609.78 & 105.02 \\
 & FlowR3-70    & 0.000 & 0.021 & 0.090 & 0.607 &  566.05 & 126.63 \\
 & FlowR3-80    & 0.000 & 0.026 & 0.065 & 0.618 &  513.13 & 122.45 \\
 & RFM-70       & 0.000 & 0.042 & 0.111 & 0.582 &  643.31 & 158.68 \\
 & RFM-80       & 0.000 & \textbf{0.043} & \textbf{0.113} & \textbf{0.594} & \textbf{513.13} & \textbf{105.02} \\
\hline
\end{tabular}
\end{adjustbox}
\end{table*}

\begin{table*}[ht]
\centering
\caption{Zero-shot performance of RFM-70 and RFM-80 models on the MultiIAN dataset across different pre-training datasets.}
\label{tab:genbaseline_ian}
\begin{adjustbox}{width=0.9\linewidth}
\begin{tabular}{l l ccccccc}
\toprule
Pre-training & Method & Acc@1km & Acc@25km & Acc@50km & Acc@200km & MeanDist (km) & MedianDist (km) \\
\midrule
\multirow{2}{*}{iNaturalist}
 & RFM-70 & 0.000 & 0.004 & 0.027 & 0.291 & 424.40 & 1213.17 \\
 & RFM-80 & 0.000 & 0.010 & 0.038 & 0.291  & \textbf{420.69} & \textbf{1199.87} \\
\midrule
\multirow{2}{*}{YFCC}
 & RFM-70 & 0.001 & 0.019 & 0.077 & 0.666 & 125.76 & 654.83 \\
 & RFM-80 & 0.000 & 0.026 & 0.084 & 0.709 & \textbf{117.98} & \textbf{623.79} \\
\midrule
\multirow{2}{*}{OSV-5M}
 & RFM-70 & 0.000 & 0.034 & 0.109 & 0.746 & 103.91 & 169.60 \\
 & RFM-80 & 0.000 & 0.032 & 0.094 & 0.741 & \textbf{99.93} & \textbf{174.55} \\
\bottomrule
\end{tabular}
\end{adjustbox}
\end{table*}

\textbf{Generative baselines:} Table \ref{tab:genbaseline_multi} and \ref{tab:genbaseline_ian} provide baseline performance of three different variants of ProbGLC (e.g., Diffusion, Flow Matching, and Riemannian Flow Matching) across the SAGINDisaster and MultiIAN dataset, respectively. 

Regarding the ProbGLC pre-training options, those baseline models pre-trained on YFCC and OSV-5M consistently lead to higher accuracy and lower localization errors compared to iNaturalist. In the SAGINDisaster dataset (Table \ref{tab:genbaseline_multi}), RFM-80 achieves Acc@50km of 0.101 with Mean/Median Dist of 941.71 km and 165.56 km on YFCC, and Acc@50km of 0.113 with Mean/Median Dist of 513.13 km and 105.02 km on OSV-5M. In contrast, baseline models pre-trained on iNaturalist perform poorly, with Acc@50km below 0.04 and Mean Dist exceeding 1760 km, even Diffusion-70 only reaches Acc@200km of 0.269, and FlowR3-70 has Mean Dist of 2493.61 km. As for the MultiIAN dataset (Table \ref{tab:genbaseline_ian}), a similar trend is observed. RFM-80 reaches Acc@50km of 0.477 on OSV-5M with Mean/Median Dist of 99.93 km and 174.55 km, while on YFCC it achieves Acc@50km of 0.291 with Mean/Median Dist of 117.98 km and 623.79 km. iNaturalist-pre-trained models have lower accuracies, with Acc@50km below 0.23 and median distances exceeding 1199 km. Notice, for Acc@1km, all baseline models despite of pre-training and test split, failed almost completely in the SAGINDisaster dataset, and achieved very poor accuracy even in the local-scale MultiIAN dataset. This observation highlights the substantial gap when introducing generative geolocalization approaches to the critical disaster response scenario.

Furthermore, two important findings herein deserve attention, thus affecting the upcoming evaluation design: 1) since the RFM consistently outperforms the other two variants, we decided to use it as the default training strategy in the ProbGLC. Similarly, we selected the YFCC and OSM-5M as pre-training baselines for further comparison given their performance superiority. 2) The key advantage of the generative approach lies in the fact that it directly predicts real locations of VGI imagery on the Earth rather than matching it with RSI to decide the location. 

As the generative approach generates a probabilistic prediction of the imagery location, this brings a unique advantage for disaster geolocalization in terms of model explainability and uncertainty quantification, besides the potential of performance boosting. In the next section, we will elaborate on the performance gain as well as the uncertainty quantification of the ProbGLC as a combined approach of both deterministic and generative geolocalization methods.

\begin{table*}[ht]
\centering
\caption{Fine-tuning results on the SAGINDisaster dataset with a 50 km threshold, showing the impact of refinement and retrieval. Herein, RFM refer to standalone generative model of Riemannian Flow Matching, Retrieval is the deterministic cross-view imagery retrieval baseline, and ProbGLC is the proposed approach. Best results for each pre-trained dataset are in bold.}
\label{tab:probglc_multi}
\begin{adjustbox}{width=\linewidth}
\begin{tabular}{llccccccc}
\toprule
Pre-trained & Method & Fine-tuned & Acc@1km & Acc@25km & Acc@50km & Acc@200km & Mean Dist (km) & Median Dist (km) \\
\midrule
YFCC
 & RFM-70  & \xmark & 0.006 & 0.054 & 0.096 & 0.537 & 1078.54 & 187.37 \\
 & RFM-80  & \xmark & 0.014 & 0.055 & 0.101 & 0.558 & 941.71 & 165.56 \\
 & RFM-70         & \checkmark & 0.000 & 0.646 & 0.681 & 0.784 & 273.97 & 7.68 \\
 & RFM-80         & \checkmark & 0.000 & 0.625 & 0.666 & 0.796 & 267.58 & 7.71 \\
 & Retrieval@1-70 & \checkmark & 0.446 & 0.678 & 0.692 & 0.753 & 222.59 & 0.94 \\
 & Retrieval@1-80 & \checkmark & 0.459 & 0.668 & 0.685 & 0.750 & \textbf{154.05} & \textbf{0.86} \\
 & ProbGLC-70     & \checkmark & 0.446 & 0.678 & 0.692 & 0.787 & 268.85 & 1.31 \\
 & ProbGLC-80     & \checkmark & \textbf{0.459} & \textbf{0.668} & \textbf{0.685} & \textbf{0.796} & 261.62 & 1.32 \\
\midrule
OSM-5M
 & RFM-70   & \xmark     & 0.000 & 0.042 & 0.111 & 0.582 &  643.31 & 158.68 \\
 & RFM-80   & \xmark    & 0.000 & 0.043 & 0.113 & 0.594 & 513.13 & 105.02 \\
 & RFM-70         & \checkmark & 0.000 & 0.638 & 0.710 & 0.859 & 142.32 & 12.24 \\
 & RFM-80         & \checkmark & 0.000 & 0.570 & 0.680 & 0.844 & 165.31 & 14.40 \\
 & Retrieval@1-70 & \checkmark & 0.468 & 0.716 & 0.736 & 0.801 & 72.91 & 0.86 \\
 & Retrieval@1-80 & \checkmark & 0.469 & 0.683 & 0.707 & 0.762 & \textbf{70.74} & \textbf{0.77} \\
 & ProbGLC-70     & \checkmark & 0.468 & 0.716 & 0.736 & 0.861 & 133.52 & 1.18 \\
 & ProbGLC-80     & \checkmark & \textbf{0.469} & \textbf{0.683} & \textbf{0.707} & \textbf{0.849} & 156.25 & 1.27 \\
\bottomrule
\end{tabular}
\end{adjustbox}
\end{table*}

\begin{figure*}[!t]
\centering
\centering
\includegraphics[width=\textwidth]{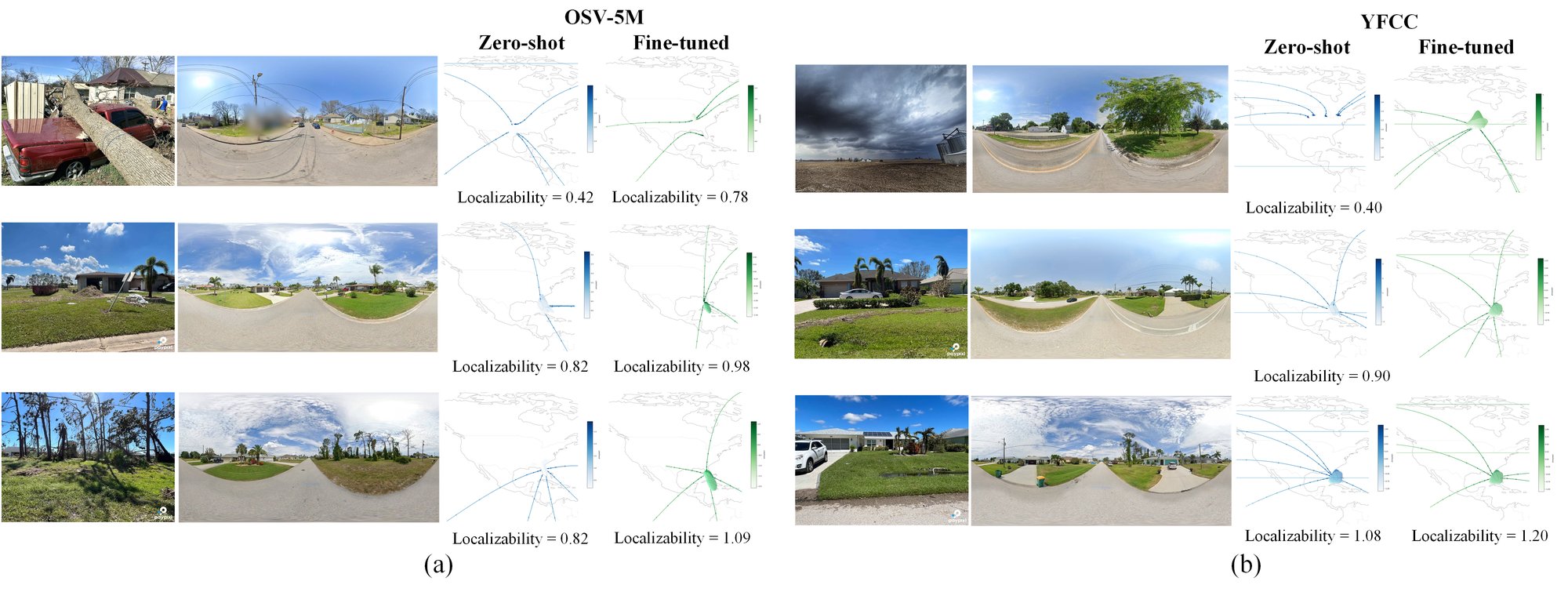}

\caption{Examples of VGI imagery and corresponding SVI, and the estimation of localizability under different training strategies for the SAGINDisaster dataset. (a) Results based on OSV-5M pre-training; (b) Results based on YFCC pre-training.}
\label{fig:Localizability_multi}
\end{figure*}

\begin{table*}[ht]
\centering
\caption{MultiIAN dataset: Fine-tuning with 50km threshold and impact of refinement and retrieval. Herein, RFM refer to standalone generative model of Riemannian Flow Matching, Retrieval is the deterministic cross-view imagery retrieval baseline, and ProbGLC is the proposed approach. Best results per pre-trained dataset are highlighted.}
\label{tab:probglc_ian}
\begin{adjustbox}{width=1\linewidth}
\begin{tabular}{llccccccc}
\toprule
Pre-trained & Method & Fine-tuned & Acc@1km & Acc@25km & Acc@50km & Acc@200km & Mean Dist (km) & Median Dist (km) \\
\midrule
YFCC
 & RFM-70 & \xmark & 0.001 & 0.019 & 0.077 & 0.666 & 125.76 & 654.83 \\
 & RFM-80 & \xmark & 0.000 & 0.026 & 0.084 & 0.709 & 117.98 & 623.79 \\
 & RFM-70         & \checkmark & 0.114 & 0.949 & 0.963 & 0.987 & 9.88 & 2.50 \\
 & RFM-80         & \checkmark & 0.090 & 0.952 & 0.968 & 0.994 & 8.01 & 2.92 \\
 & Retrieval@1-70 & \checkmark & 0.838 & 0.962 & 0.970 & 0.981 & 4.25 & 0.00 \\
 & Retrieval@1-80 & \checkmark & 0.868 & 0.970 & 0.977 & 0.994 & \textbf{3.85} & 0.00 \\
 & ProbGLC-70     & \checkmark & 0.838 & 0.962 & 0.970 & 0.986 & 6.77 & 0.00 \\
 & ProbGLC-80     & \checkmark & \textbf{0.868} & \textbf{0.970} & \textbf{0.977} & \textbf{0.996} & 4.00 & \textbf{0.00} \\
\midrule
OSV-5M
 & RFM-70 & \xmark & 0.000 & 0.034 & 0.109 & 0.746 & 103.91 & 169.60 \\
 & RFM-80 & \xmark & 0.000 & 0.032 & 0.094 & 0.741 & 99.93 & 174.55 \\
 & RFM-70         & \checkmark & 0.085 & 0.957 & 0.973 & 0.992 & 9.09 & 3.12 \\
 & RFM-80         & \checkmark & 0.100 & 0.963 & 0.975 & 0.994 & 7.48 & 2.86 \\
 & Retrieval@1-70 & \checkmark & 0.848 & 0.970 & 0.974 & 0.987 & 3.55 & 0.00 \\
 & Retrieval@1-80 & \checkmark & 0.870 & 0.972 & 0.978 & 0.991 & \textbf{3.09} & 0.00 \\
 & ProbGLC-70     & \checkmark & 0.848 & 0.970 & 0.974 & 0.992 & 5.49 & 0.00 \\
 & ProbGLC-80     & \checkmark & \textbf{0.870} & \textbf{0.972} & \textbf{0.978} & \textbf{0.994} & 3.85 & \textbf{0.00} \\
\bottomrule
\end{tabular}
\end{adjustbox}
\end{table*}

\begin{figure*}[!t]
\centering
\centering
\includegraphics[width=\textwidth]{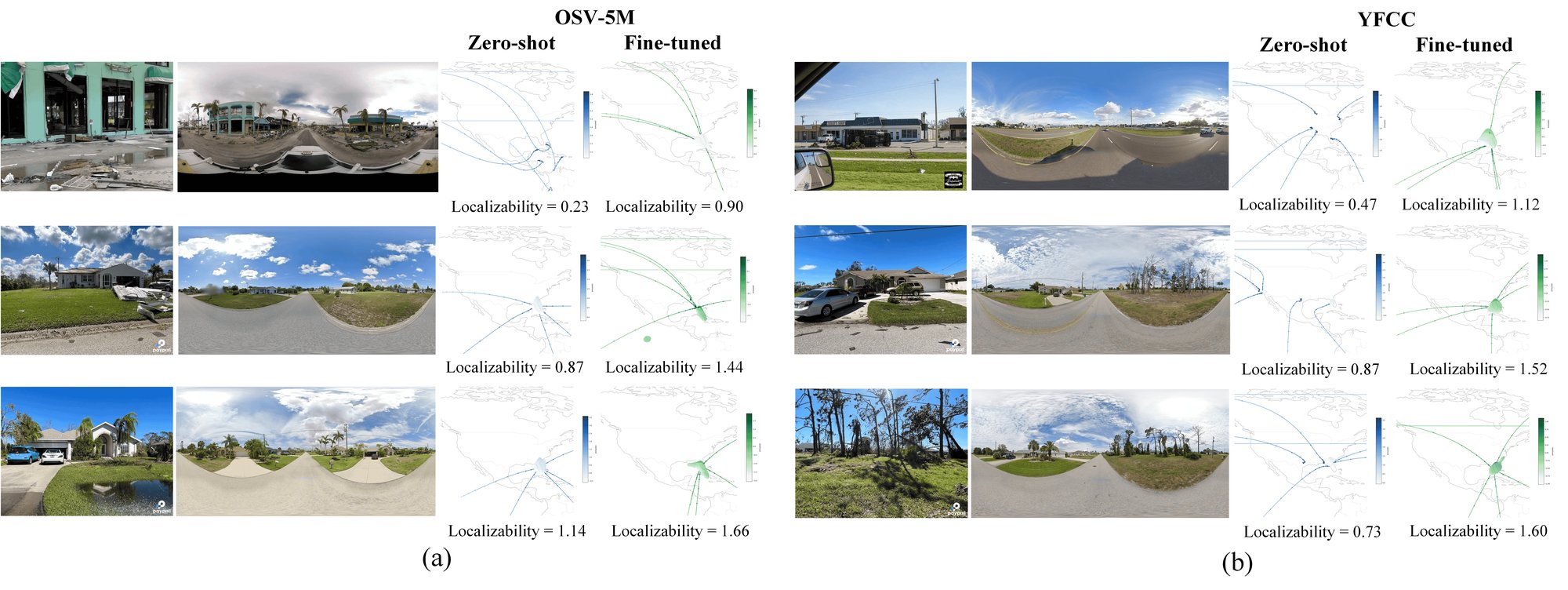}
\caption{Examples of VGI imagery and corresponding SVI, and the estimation of localizability under different training strategies for the Hurricane IAN dataset. (a) Results based on OSV-5M pre-training; (b) Results based on YFCC pre-training.}
\label{fig:Localizability_ian}
\end{figure*}

\subsection{ProbGLC Geolocalization Results}\label{probGLC result}

\textbf{Two-level of Performance Gain:} The proposed ProbGLC approach aims at two levels of improvements over 1) baseline pre-training models and 2) generative or retrieval only fine-tuning models. Table \ref{tab:probglc_multi} and \ref{tab:probglc_ian} summarize our experiment results of evaluating ProbGLC approaches (applying a threshold $\mathbf{r}$ equal to 50 km), across two disaster datasets, respectively. As a protocol, the comparison between generative-only (i.e., RFM), retrieval-only, and the ProbGLC approaches highlights 1) the performance gain over baselines, 2) the advantage of uncertainty quantification and localizability measure via probabilistic predictions. 

As one can see in Table \ref{tab:probglc_multi}, our ProbGLC led to substantial performance gains on the SAGINDisaster dataset across both YFCC and OSV-5M pre-trained baseline models. Compared to the pre-training baseline, where Acc@50km remained below 0.11 (e.g., 0.096 for YFCC with RFM-70 and 0.111 for OSM-5M with RFM-70), the ProbGLC boosted it to around 0.7 in Acc@50km. This first-level improvement is accompanied by a major reduction in Mean Dist, from over 1000 km to about 270 km for YFCC and 140 km for OSM-5M, and a sharp decrease in Median Dist from more than 150 km to around 7–12 km, which is favorable toward closing the "last kilometer" challenge in a disaster response scenario. Moreover, the ProbGLC variants as well as the deterministic Retrieval@1 notably enhanced short-range accuracy, especially for reaching up to 0.469 for Acc@1km and around 0.736 for Acc@25km, reducing the Median Dist to sub-kilometer levels (0.86 km for OSV-5M and 0.86–1.32 km for YFCC). However, we want to emphasize again that the key advantage of ProbGLC over deterministic Retrieval@1 is not about superior performance as they both yield very competitive results, but more about the explainability feature that comes with the generative probabilistic geolocalization approach.

Among all settings, OSM-5M consistently achieved lower Mean Dist (down to 70.74 km) and higher Acc@25km to Acc@200km, suggesting stronger spatial generalization.  In contrast, YFCC exhibited slightly larger residual Mean Dist despite comparable Acc@50km, indicating a slightly weaker localization accuracy. We attribute this phenomena to the fact that OSM-5M focuses on street-level features, which present higher similarity to the disaster-related VGI imagery in the latent space. Overall, we can see that ProbGLC is able to simultaneously benefit from the fine-tuning of generative approach and the deterministic refinement of retrieval-based approach, producing the best balance between accuracy and distance metrics across all scales.

Compared with the SAGINDisaster results (Table \ref{tab:probglc_multi}), the fine-tuning on the MultiIAN dataset led to dramatically higher localization accuracy and lower distance errors as shown in Table \ref{tab:probglc_ian}. For ProbGLC variants, Acc@50km rose from roughly 0.68–0.71 to above 0.96 for both YFCC and OSV-5M models, while the Mean Dist dropped sharply from over 270 km to under 10 km. The combined ProbGLC approach further boosted short-range accuracy, achieving up to 0.870 at 1 km and reducing the Median Dist to 0.00 km, indicating near-perfect geolocalization. This is intuitive as we already mentioned that SAGINDisaster is considered in general more challenging than MultiIAN. However, the implication on a local-scale disaster dataset is indeed exciting as it means by incorporating local samples, the ProbGLC approach can easily boost up a very competitive approach which is crucial when deploying to new disaster on a timely basis.

\textbf{Probabilistic Distribution and Localizability:} As aforementioned, the unique advantage of generative approach (with RFM) is about the predicted location comes always with a probabilistic distribution as well as their fixed velocity field as illustrated in Figure \ref{fig:RFM}, which enables us to visualize the conferencing trajectory of different VGI imagery directly in the Earth as well as calculating their localizability score as a uncertainty proxy. 

Figure \ref{fig:Localizability_multi} and \ref{fig:Localizability_ian} offer a visual-intuitive comparison of generative RFM trajectories for baseline models and the fine-tuned ProbGLC approach alongside the corresponding localizability scores as a major uncertainty metric. First, we see a clear improvement of the ProGLC over generative baselines as the probability distributions are more spatially clustered as well as the increase localizability scores. In this context, by interpreting the trajectory patterns, one can now better explain where exactly the predicted locations and how confident the models are, which provides key insights into designing more focused and uncertainty-aware disaster response actions. Second, we notice a common character of high localizability scores still refers to street layout (e.g., pavement and cross-road) as well as landscape features (e.g., palm tree and building styles), which is helpful to guide the further development in global-scale imagery geolocalization pre-training and how to best deploy the ProbGLC approach in future disaster events. The capability of understanding how and why a model predicts a location is one of the most desired features and a persistent limitation for deterministic cross-view geolocalization approaches.

\begin{figure*}[t]
    \centering
    \begin{subfigure}[b]{0.48\textwidth}
        \centering
        \includegraphics[width=\textwidth]{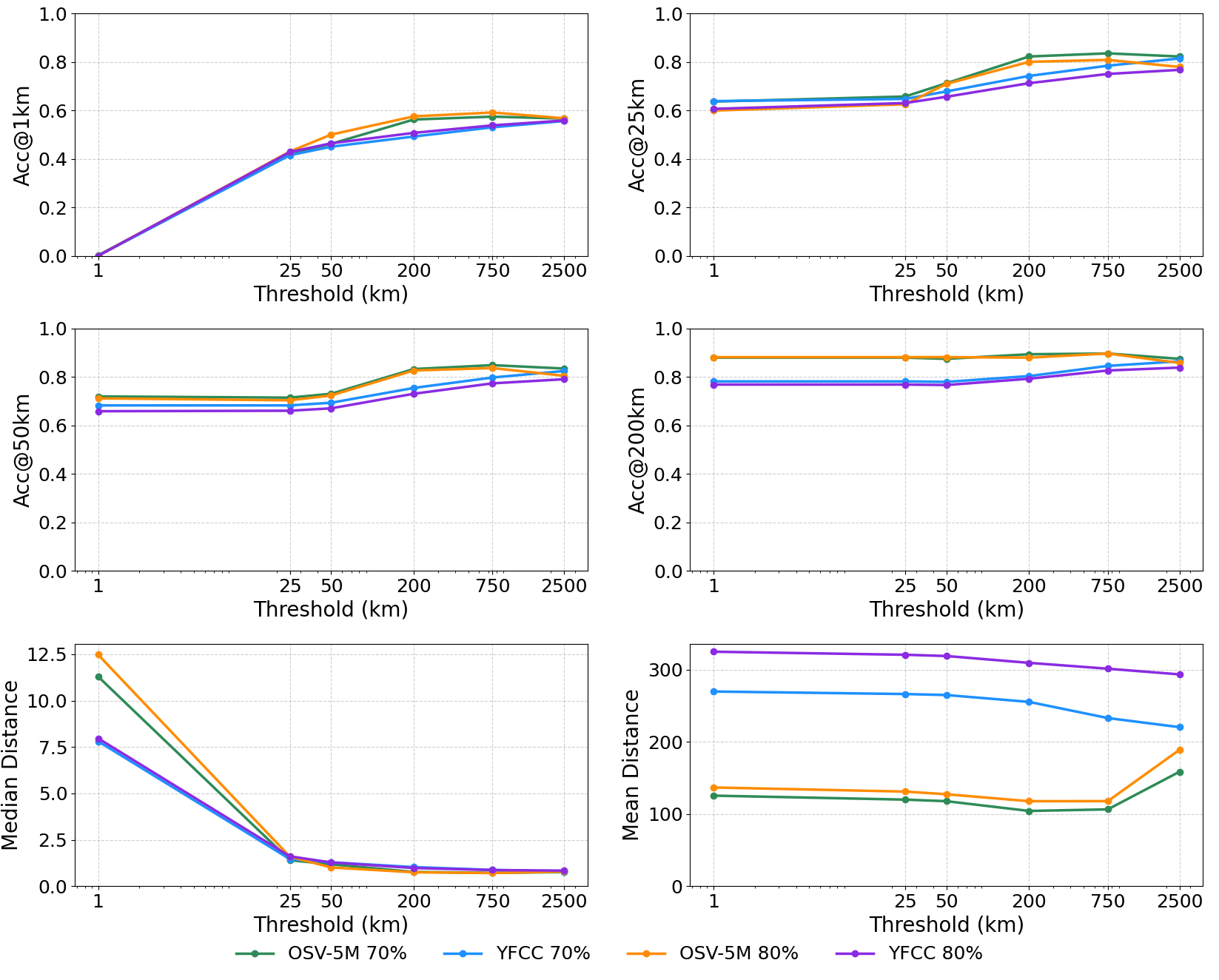}
        \caption{SAGINDisaster}
        \label{Table:Final_multi}
    \end{subfigure}
    \hfill
    \begin{subfigure}[b]{0.48\textwidth}
        \centering
        \includegraphics[width=\textwidth]{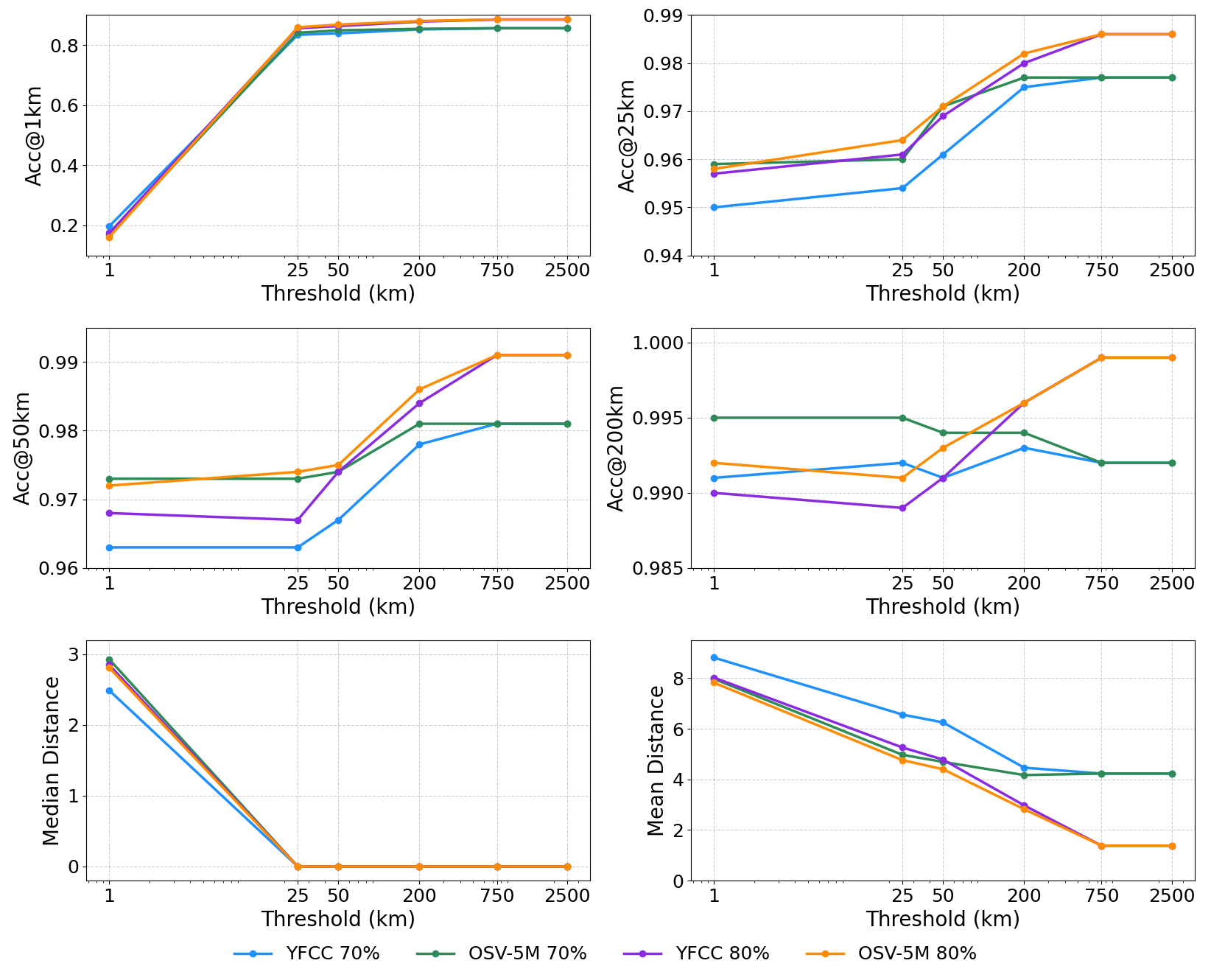}
        \caption{MultiIAN}
        \label{Table:Final_IAN}
    \end{subfigure}
    \caption{Effect of different decision thresholds on OSV-5M and YFCC across two datasets.}
    \label{fig:both_datasets}
\end{figure*}

\textbf{Ablation Study:} To provide a systematic justification, we examine the impact of different threshold $\mathbf{r}$, which is considered as a trade-off between narrowing the probabilistic searching zone and maximizing deterministic retrieval accuracy. To remind its function, the threshold $\mathbf{r}$ determines the extent to which the Retrieval@1 models consider the predicted locations from the RFM generative models. Figure \ref{Table:Final_multi} and \ref{Table:Final_IAN} elaborate on the impact of different distance thresholds on the SAGINDisaster and MultiIAN datasets, respectively. Two common patterns deserve notice here: 1) Median and Mean distances drop dramatically already when relaxing the threshold from 1km to 25km, aligning with the steep increase in fine-grained accuracy (Acc@1km and Acc@25km). This observation confirms the effectiveness of the combined ProbGLC approach to simultaneously benefit from the generative and deterministic geolocalization approaches, ensuring both model explainability and accuracy performances. 2) One can notice that the change of threshold doesn't influence the performance much when the distance is set bigger than 50km, which actually means that the largest performance gain comes with the fact that the generative approach is able to narrow down the location to a relatively small area where the deterministic approach is able to hit the exact location via imagery retrieval. This is also the reason why we set the threshold $\mathbf{r}$ to 50km in Table \ref{tab:probglc_multi} and \ref{tab:probglc_ian}. Of course, this ablation is not supposed to be comprehensive as one can still try out different settings in both RFM or Retrieval@1 approaches. However, as aforementioned, the threshold $\mathbf{r}$ is considered a key hyperparameter of the ProbGLC approach to balance accuracy and explainability from either deterministic or probabilistic approaches. If we take ProbGLC as a general framework for cross-view geolocalization, one could easily extend this ablation to other important parameters. 

\begin{figure*}[!t]
\centering
\centering
\includegraphics[width=\textwidth]{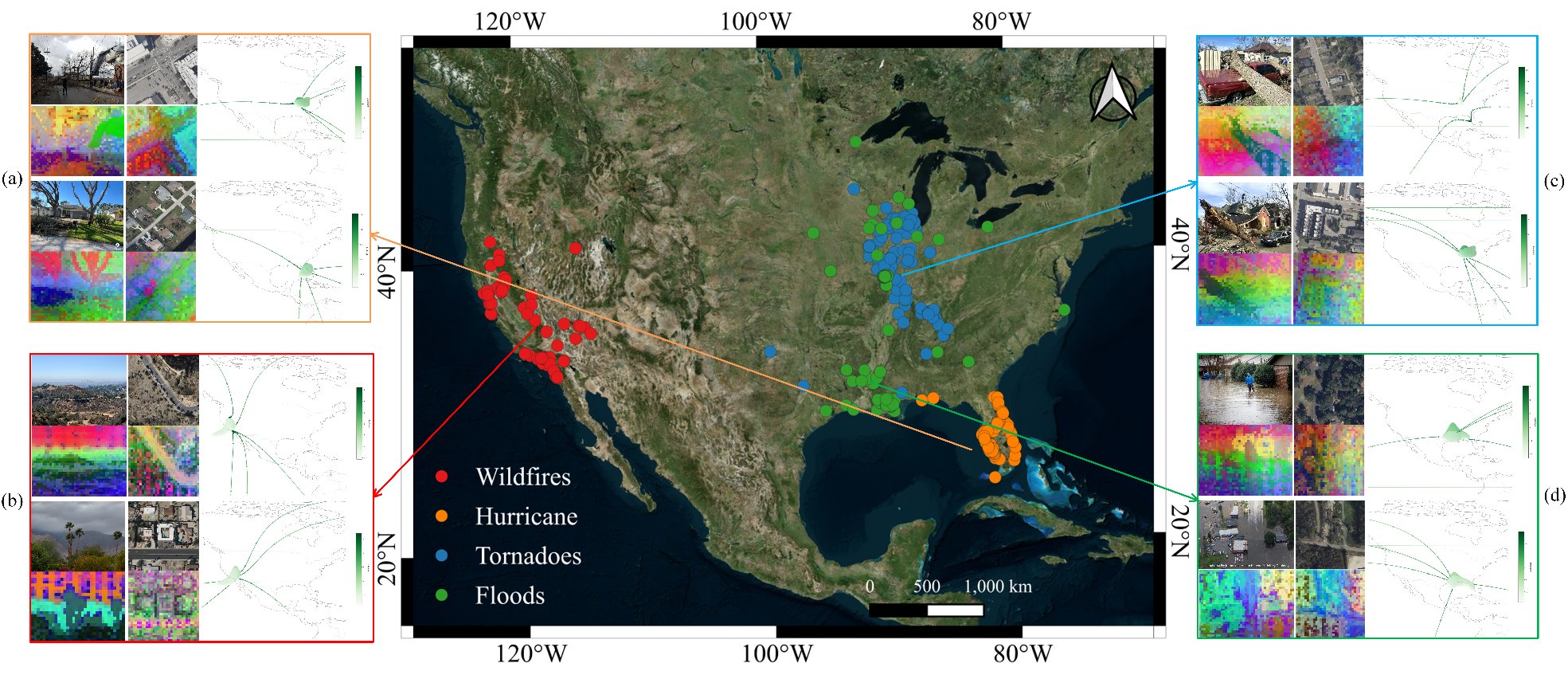}
\caption{Visualization results of several representative examples from the SAGINDisaster dataset, corresponding to four major disaster types: (a) Hurricane, (b) Wildfire, (c) Tornado, and (d) Flood.}
\label{fig:Final_multi}
\end{figure*}

\subsection{Implications in Disaster Response}\label{disaster result}

In this section, we would like to elaborate on the implications in real-world disaster response scenarios that we foresee for the proposed ProbGLC. 

First, as shown in Figure \ref{fig:Localizability_multi} and \ref{fig:Localizability_ian}, the localizability score together with the probabilistic density distribution is key to prioritize resource allocation. For example, decision-makers can use the localizability score to filter real-time VGI disaster imagery. Images with high localizability (low uncertainty) can be geolocated and immediately included into a crisis map for rapid damage assessment, while images with low localizability (high uncertainty) often indicate difficult terrains or severe destruction where visual cues are lost. These specific cases can be routed to human analysts for expert review, or directing UAVs to the estimated general area to capture clearer imagery where ground-level ambiguity is too high, preventing the dispatch of rescue teams to the wrong location.  

Second, the ProbGLC allows for optimizing searching zones in real disaster response. For instance, in case of a lost person or a trapped group identified via real-time VGI imagery, the density map defines the geographical extent for the search operation. A tight cluster implies a focused search radius (e.g., Acc@1km), while a dispersed cloud suggests a wider search grid is necessary. By combining with the localizability score, the whole system allows decision-makers to calibrate their confidence level in real-time and optimize their response actions with more informed decisions.

Moreover, thanks to the deterministic image encoder (i.e., ViT DINOv2), we are now able to visualize and compare the learned cross-view image patterns regarding different disaster types. Herein, Figure \ref{fig:Final_multi} presents a qualitative visual comparison using DINOv2 PCA techniques \citep{dinov2github} with selective examples from four major types of disaster in the SAGINDisaster dataset, respectively. Herein, DINOv2 PCA Visualization does the following trick, we have an image, process it through the Transformer and then have 32x32 patches. All have their features in the dimension size of the Transformer (i.e., 1024 in the case of ViT-L). Then we fit a PCA to 3 components meaning instead of 32x32x1024 we get 32x32x3. This is very intriguing because we can now interpret the 3 dimensions as colors, namely RGB. As we can see in Figure \ref{fig:Final_multi} when we compare PCA of VGI imagery the features tend to have the same color through the same PCA. For instance, the water is blue in (d), the tree and vegetation are purple (c), and the sky is orange (b). In the RSI image, there are more colors as more is visible compared to the VGI imagery. But this can be very helpful because it means the learned cross-view embeddings for the ProbGLC approach encode direct semantic relations, which is what is in the image and what is very similar. When combined with the probabilistic density map, this kind of visualization offers good insights into common and distinct features in different disaster-related geolocalization tasks.

\section{Discussions} \label{dicussion}
In this work, we present a \textbf{Prob}abilistic Cross-view \textbf{G}eo\textbf{L}o\textbf{C}alization approach, in short \textbf{ProbGLC}, as a novel pathway towards generative location awareness for rapid disaster response. Despite the motivating preliminary results we got, there are a few points that deserve additional discussion and awareness. 

\textbf{Generative cross-view approaches:} A major contribution of the proposed ProbGLC approach is to introduce the generative training approach (via diffusion and flow matching) into imagery geolocalization tasks. However, the pretraining of these generative backbones (e.g., Diffusion, FlowR3, RFM) is limited largely by two practical factors: 1) the geographical distribution, density, and focus of three global visual geolocalization datasets, namely YFCC, OSV-5M, and iNaturalist; 2) the sole training target of locations (i.e., lat and long). The first factor implies the underlying biases in the crowdsourced imagery datasets, where the density and spatial coverage of images, no matter street view or social media, vary significantly across countries and regions. Therefore, the geographical generalizability \citep{li2023rethink}, similarly the replicability across space and time \citep{goodchild2021replication}, deserves more dedicated and comprehensive evaluations and treatments. The second factor can be attributed to the fact that there is still a research gap in a global-scale geospatial pretraining scheme that can provide fine-grained and general-purpose backbones. Recent works, including but not limited to  SatCLIP \citep{klemmer2025satclip}, Clay Foundation Model \citep{clay2024huggingface}, and the most recent Google AlphaEarth Embedding \citep{brown2025alphaearth}, shed light on this direction, while their performance in time-critical disaster response scenarios still needs careful investigations and the potential of generative cross-view training remains largely unexplored. 

\textbf{GeoAI for disaster resilience:} In this work, we focus on ensuring fast and accurate cross-view geolocalization in a post-disaster scenario. Preliminary results from diverse disaster types (i.e., hurricane, wildfire, tornado, and flood) highlighted the increasing role of GeoAI to predict, analyze, and enhance urban disaster resilience against climate extreme events. A comprehensive assessment of disaster resilience involves not only physical infrastructure such as drainage systems and flood barriers, but also socio-economic factors like population density, vulnerability, and impact assessment.

Towards multi-temporal monitoring and fine-grained disaster mapping, we wish the lessons learned from this work can be integrated into dedicated local GeoAI initiatives, especially for those global south areas as argued in \citep{huang2025geospatial}, to better prepare our cities for future disaster and climate extreme events. Recent works by \citet{xue2025regression, wang2026cityvlm} highlight the potential of incorporating Vision-Language Models (VLMs) in understanding fine-grained and regression-like geospatial predictions. It would be interesting to see how one can leverage VLMs to understand the multi-faceted disaster resilience aspects (before, during, and after) in combination with socio-economic context information. 

\textbf{Representation learning for multimodal geospatial data:} A fundamental task for GeoAI models is to align the multimodal geospatial data in the latent space so that their latent representations can be better used for downstream tasks \citep{ mai2024srl,liu2025representation}, no matter whether it is regression, classification, or geolocalization tasks. However, so far the major advances in the field focus on raster-based imagery data (such as satellite and street view imagery) due to the uniform format of pixels and rich content from the visual perception. However, in the geospatial domain, there is still a large amount of vector data (e.g., points, polylines, polygons) that remains largely underexplored for representation learning, for instance, in OpenStreetMap, which also contains rich and helpful location and semantic information. 

To extract meaningful features from vector data for predictive tasks, existing researchers either perform feature engineering based on domain knowledge \citep{yan2022graph}, or convert spatial data from their original formats (e.g., points, polylines, and polygons) into formats that are easier for neural networks to handle, or map vector files into raster image tiles \citep{kang2019transferring, li2022conceptual}. Herein, the first stream heavily relies on domain knowledge and can not be easily generalized to new tasks, while the latter stream suffers from reduced data precision and loss of linked semantic context. Either way, the representation capability and performance are limited due to the lack of end-to-end learning. Therefore, a promising direction for future work is to incorporate massive vector datasets (e.g., from OpenStreetMap) into the current learning framework with a dedicated vector data encoder, for instance, Poly2Vec \citep{siampou2024poly2vec} and Geo2Vec \citep{chu2025geo2vec}, facilitating a synergy of multimodal geospatial data into more efficient and effective representation learning.

\section{Conclusions} \label{conclusion}

In this paper, we present a novel cross-view disaster geolocalization approach, called ProbGLC, standing for \textbf{Prob}abilistic Cross-view \textbf{G}eo\textbf{L}o\textbf{C}alization, to facilitate generative location awareness for rapid disaster response purpose. The ProbGLC is the first of this kind of model to combine probabilistic and deterministic geolocalization models into a unified framework, which brings unique benefits in terms of model explainability (via uncertainty quantification) and superior geolocalization accuracy compared to state-of-the-art baseline approaches. More importantly, by leveraging the generative training strategy, for instance, Riemannian Flow Matching on the Sphere, the ProbGLC predicts imagery locations directly on the Sphere with a probabilistic distribution as a proxy of uncertainty and localizability, which enables us to better understand how the model makes decisions during the geolocalization process. Extensive experiments on two cross-view disaster geolocalization datasets (SAGINDisaster and MultiIAN) show competitive performance and explainability features of the proposed ProbGLC model in handling multiple disaster types with distinct disaster characteristics. In addition, we carefully elaborate on the implications of ProbGLC in various disaster response scenarios and the benefits it brings towards leveraging GeoAI in disaster response. Building on these insights, our future work will focus on developing responsible GeoAI approaches that enhance disaster risk management and societal resilience in the changing climate.

\section*{Acknowledgments} 

We would like to thanks the editor and reviewers for their valuable comments, which have helped improve the quality of our manuscript. This work was supported by the Start-Up Grant (SUG) project “Geospatial Artificial Intelligence for Climate Resilient Urban Environment” from the National University of Singapore (E-109-00-0036-01).

\section*{Disclosure Statement} No potential conflict of interest was reported by the authors.

\appendix
{
\color{black}

\section{Geographic Diffusion in the Euclidean space}\label{Appen_A}

In this appendix, we elaborate on the basic of geographic diffusion appraoch in the Euclidean space. Since this work is among the early attempts in the geospatial community to address imagery geolocalization from a generative perspective, we think it would be beneficial to provide an introduction to the basics of diffusion models. The concept of diffusion models was first introduced by \citet{DicksteinW2015}, which only gained popularity after \citet{Ho2020} formally outlined their use as generative models.

\textbf{Forward Diffusion Process}: Given an input variable, in our case a random location $x_n$ ( longitude and latitude), the core idea of diffusion is to iteratively add Gaussian noise to this location variable. We denote the location at the $0$th timestamp (i.e., the original location where no noise has been added yet) as $\mathbf{x}_0$. With each iteration, we increment the timestamp such that $\mathbf{x}_t = \mathbf{x}_{t-1} + \epsilon$, where $\epsilon \in \mathbb{R}^D$ refer to the Gaussian noise. After a sufficient number of iterations, we posit that $\textbf{x}_T \sim \mathcal{N}(0, \mathbf{I})$. In other words, if we add enough noise, the resulting location variable will be indistinguishable from a random one drawn from a unit normal distribution. This process is called the forward diffusion process, which can be technically formulated as follows:

\begin{equation}\label{eq1}
    \mathbf{x}_t = \sqrt{1 - \beta(t)}\mathbf{x}_{t-1} + \sqrt{\beta(t)}\epsilon,
\end{equation}

Where $\mathbf{x}_1,..., \mathbf{x}_T$ are the noisy latent variables at different time stamps $t \in [0, 1]$ and $\beta(t) : [0, 1] \to [0, 1]$ as a scheduling function with $\beta(t) = 0$ and $\beta(t) = 1$ to control the noise level added to the coordinates.

Herein, these latent variables are not simply fixed versions of the noisy $x_0$, but a condition variable. We assume that they follow a distribution conditioned on the previous step, denoted as $\mathcal{N}(0, \beta(t)\mathbf{I})$ that represents a multivariate Gaussian distribution with covariance of $\beta(t)\mathbf{I}$. The only exception will be the original location, denoted as $\mathbf{x}_0 \sim q(x_0)$, which is not conditioned on anything else. Therefore, the sequence of diffused locations $\mathbf{x}_1$ to $\mathbf{x}_T$ forms a Markov chain and can be formulated as:

\begin{equation} \label{eq2}
   q(\mathbf{x}_t|\mathbf{x}_{t-1}) = \mathcal{N}(\sqrt{1 - \beta(t)}\mathbf{x}_{t-1}, \beta(t)\mathbf{I}),
\end{equation}

Which further leads to the following equation to present the whole forward diffusion process:

\begin{equation} \label{eq3}
    q(\mathbf{x}_{1:T}|\mathbf{x}_0) = \prod_{i=1}^T q(\mathbf{x}_t|\mathbf{x}_{t-1}), 
\end{equation}

\textbf{Backward Denoising Process}: Based on the forward diffusion process, the key idea of Denoising Diffusion Probabilistic Models (DDPM) introduced by \citet{Ho2020} is to learn a neural network parametrized by $\mathbb{\varphi}$, which takes the diffused location $\mathbf{x}_t$ and the noisy level $\beta(t)$ as input and most importantly is conditioned on the imagery embeddings $\mathbf{c}$.

Considering the final timestep $T$ of added noise will result in an entirely random location. We can represent this as $P(\mathbf{x}_T) = \mathcal{N}(0, \mathbf{I})$, where $P_\mathbb{\varphi}(\mathbf{x}_{t-1}|\mathbf{x}_t)$ denotes the learned distribution of the de-noised $\mathbf{x}_t$. This represents the previous step in the noise addition process, $\mathbf{x}_{t-1}$. Just opposite to the forward diffusion process, the backward denosing will yield:

\begin{equation} \label{eq4}
    P_\mathbb{\varphi}(\mathbf{x}_{t-1}|\mathbf{x}_t) = \mathcal{N}(\mu_\mathbb{\varphi}(\mathbf{x}_t, t), \Sigma_\mathbb{\varphi}(\mathbf{x}_t, t)),
\end{equation}

Where $\mu_\mathbb{\varphi}$ and $\Sigma_\mathbb{\varphi}$ refer to the mean and covariance nosiy vector $\epsilon$ for the previous state $\mathbf{x}_{t-1}$, both are functions of $\mathbf{x}_t$ and the timestamp $t$. Herein, $\mathbb{\varphi}$ refers to a set of trainable parameters in the neural network (Figure \ref{fig:task_statement}), with which we can learn a function to de-noise the latent variables.

The overall denoising process can then be formulated similarly as a Markov chain as follows:

\begin{equation} \label{eq5}
    P_\mathbb{\varphi}(\mathbf{x}_{0:T}) = P(\mathbf{x}_T) \prod_{i=1}^T P_\mathbb{\varphi}(\mathbf{x}_{t-1}|\mathbf{x}_t),
\end{equation}

The key is to find a solution to optimize $\mathbb{\varphi}$ to achieve a one-to-one denoising of the locations. As introduced by \citet{Ho2020}, it is suggested to fix the variance to 1, so the only remaining learnable parameter is $\mu_\mathbb{\varphi}(\mathbf{x}_t, t)$. Further, the $\mu_\mathbb{\varphi}$ predictor is replaced with the $\epsilon$, where the predictor is trained to predict the reverse noise, in order to denoise the latent variable. As a result, the whole generative model can be trained to minimize the diffusion loss function as follows:

\begin{equation}\label{eq6}
    \mathcal{L}_{DF} = \mathbb{E}_{\mathbf{x}_0,  \mathbf{c}, \epsilon, t} \left[ || \mathbb{\varphi}(\mathbf{x}_t \mid c) - \epsilon ||^2 \right],
\end{equation}

Where the expectation is over four variables, saying the origin location $\mathbf{x}_0$, the imagery embedding $\mathbf{c}$, the Gaussian noise $\epsilon$, and the timestamp $t$. Herein, $\mathbb{\varphi}(\mathbf{x}_t \mid c)$ refers to the generative DDPM network based on a noisy location $\mathbf{x}_t$ and conditional on the imagery embedding $\mathbf{c}$. 

\textbf{Inference Process}: For inference, given a new imagery $\mathbf{c}$, the geographic diffusion start with a random location with $\mathbf{x}_1 = \epsilon$, then iteratively denoise the location latent variable  $\mathbf{x}_t$ over a specific timesteps from $t=1$ to $t=0$, using the location variable update equation as follows:

\begin{equation} \label{eq7}
\mathbf{x}_{t - \Delta t} = \sqrt{1 - \beta(t)}\, \hat{\mathbf{x}}_t + \sqrt{\beta(t)}\, \mathbb{\varphi}(\mathbf{x}_t \mid c)
\end{equation}

\begin{equation} \label{eq8}
\hat{\mathbf{x}}_t = \frac{1}{\sqrt{1 - \beta(t)}}\, \mathbf{x}_t - \sqrt{\beta(t)}\, \mathbb{\varphi}(\mathbf{x}_t \mid c)
\end{equation}

Where $\Delta t$ and $\hat{\mathbf{x}}_t$ refer to the time step size and the denoised location variables, respectively. Finally, the $\hat{\mathbf{x}}_0$ in Euclidean space  $\mathbb{R}^3$ will be projected to the Spherical coordinates of $\mathbb{S}^2$ (i.e., longitude and latitude).

\section{The Projection between Spherical and Euclidean Spaces} \label{Appen_b}

Herein, the logarithmic map $\log_x$ maps a point $y \in \mathcal{S}_2$ onto $T_x$, the tangent space at point $x$ \citep{sommer2020introduction}:

\begin{equation}
    \log_x(y) = \frac{\theta}{\sin \theta}(y-\cos{\theta} x)~,
\end{equation}
where $\theta=\arccos(\langle x,y \rangle)$ is the angle between $x$ and $y$.

Next, the exponential map $\exp_x$ of a point $x \in \mathcal{S}_2$ maps a tangent vector $v\in T_x$ back onto th spherical space:

\begin{equation}
    \exp_x(v) = \cos(\Vert v \Vert) x + \frac{\sin( \Vert v \Vert)}{\Vert v \Vert}v~,
\end{equation}
where $\Vert v \Vert$ is the Euclidean norm of $v$.

Moreover, the projection here is fully aligned with classic map projection theory, therefore it can be potentially extend to different ellipsoid coordinates as well.

\section{Anchor-based Reranking as Postprocessing}\label{Appen_c}

 In a nutshell, the anchor-based reranking (as shwon in Figure \ref{fig:reranking}) helps to refine the initial retrieval ranking by selecting representative anchors (i.e., often the query and top results) and re-scoring candidates based on both their similarity to the query and their consistency with these anchors.

Given the learned cross-view embedding space $\R_{CV}$ and a query embedding of VGI imagery $L_c\in\mathbb{R}^d$ and candidate RSI embeddings $\{L_s^i\}_{i=1}^N$, the initial retrieval computes a base cosine similarity as follows:

\begin{equation} \label{eq21}
    S(L_c,L_i)=\frac{L_c\cdot L_s^i}{\|L_c\|\|L_s^i\|}.
\end{equation}

By selecting the top-\(k\) retrieved items as anchors $\{a_j\}_{j=1}^k$, where $a_j\in\{L_s^i\}$, we can rescore each RSI candidate by combining query–candidate similarity and anchor–candidate consistency using the following equation: 

\begin{equation} \label{eq22}
    R(L_s^i)\;=\;\alpha\,S(q,x_i)\;+\;\frac{1-\alpha}{k}\sum_{j=1}^k S(a_j,x_i),
\end{equation}

Where $\alpha\in[0,1]$ is a ratio to balance the query and anchor influence, $x_i$ and $q$ refers to the query VGI imagery and the candidate RSI embeddings, respectively. 
}

\printcredits

\clearpage

\end{document}